\documentclass{article} % For LaTeX2e
\usepackage{amsthm, amsmath, amssymb}
\usepackage[colorlinks = true, linkcolor = black, citecolor = black, final]{hyperref}
\usepackage{graphicx, multicol}
\usepackage{marvosym, wasysym}
\usepackage{fancyhdr}

\usepackage{algorithm}
\usepackage{bm}
\usepackage[noend]{algpseudocode}

\makeatletter
\renewcommand{\ALG@beginalgorithmic}{\scriptsize}
\makeatother

\usepackage[margin=1.2in]{geometry}

\algrenewcommand\algorithmicrequire{\textbf{Require:}}
\algrenewcommand\algorithmicensure{\textbf{Postcondition:}}
\usepackage{subfigure}
\usepackage{float}

\title{Competitive Multi-agent Inverse Reinforcement Learning with Sub-optimal Demonstrations}

\author{
Xingyu Wang, Diego Klabjan \\
Department of Industrial Engineering and Management Sciences \\
Northwestern University, Evanston, IL, 60208  \\
\texttt{xingyuwang2017@u.northwestern.edu, d-klabjan@northwestern.edu} \\
}

\begin{document}

\maketitle
\begin{abstract}
This paper considers the problem of inverse reinforcement learning in zero-sum stochastic games when expert demonstrations are known to be not optimal. Compared to previous works that decouple agents in the game by assuming optimality in expert strategies, we introduce a new objective function that directly pits experts against Nash Equilibrium strategies, and we design an algorithm to solve for the reward function in the context of inverse reinforcement learning with deep neural networks as model approximations. In our setting the model and algorithm do not decouple by agent. In order to find Nash Equilibrium in large-scale games, we also propose an adversarial training algorithm for zero-sum stochastic games, and show the theoretical appeal of non-existence of local optima in its objective function. In our numerical experiments, we demonstrate that our Nash Equilibrium and inverse reinforcement learning algorithms address games that are not amenable to previous approaches using tabular representations. Moreover, with sub-optimal expert demonstrations our algorithms recover both reward functions and strategies with good quality.
\end{abstract}
%\twocolumn
\section{Introduction}

In the field of reinforcement learning, various algorithms have been proposed that guide an agent to make decisions, interact with the environment, and increase the profit or return. Usually, a sequential decision-making problem is formally defined in the Markov decision process (MDP) formalism: at a given point of time the agent is subject to a certain state in the environment; different actions lead the agent to visit different states, as prescribed by a underlying state transition probability distribution that captures and describes the environment; a reward function is also associated with states (or state-action pairs) so that the agent receives immediate rewards after each decision. Reinforcement learning algorithms are reward-driven, aiming for optimal or near-optimal strategies that guide the agent to act ideally and maximize its cumulative rewards. 

As of late, multi-agent reinforcement learning, a generalization of single-agent reinforcement learning tasks, has been gaining momentum since it is aligned with the growing attention on multi-agent systems and the applications thereof. A multi-agent task can be purely competitive in its nature: most sport games are zero-sum and fall into this category; while some games or tasks can be completely cooperative, such as coordinating multiple robots to carry and transport cargo collaboratively in the shortest possible time. Moreover, both the competitive and cooperative elements can exist at the same time, most likely in cooperative tasks where each agent also bears another motivation to save its own energy or maximize its own interests.

Either for solving a single-agent task or a multi-agent system, reinforcement learning entails the knowledge of the reward function, or at least observations of immediate reward. For some learning tasks, however, we have very little or no knowledge of the reward function, but we do have access to demonstrations performed by ``experts'' in the task. For example, successful training of an auto-pilot system often takes full advantage of human driving behavior datasets. On the contrary, a naive, manually crafted artificial reward function may fail to capture the sophisticated balancing between safety, speed, and simplicity in maneuvering when driving a vehicle. 

As one of the widely used approaches to these tasks, imitation learning concentrates on recovering strategies from available demonstrations, which is justifiable as long as we believe that when experts are performing well enough, agents should be able to perform nicely by simply imitating experts' moves. An approach, termed as inverse reinforcement learning (IRL), formulates the task as solving an inverse problem, and strives to infer the reward function from demonstrations of experts. In general, the former approach is perceived as simpler, while the latter compresses task-related information into a succinct reward function that is more transferable to further applications. 

Existing IRL methodologies in competitive multi-agent IRL settings implicitly assume that demonstrations are generated by an optimal strategy, and use this optimality of expert strategies to solve for the hidden reward function. (In a multi-agent system, optimality is usually defined in the sense of Nash Equilibrium.) In such settings, the model can be decoupled into several single-agent sub-problems. Yet the more complicated the game is, the farther this optimality assumption may diverge from the truth. For instance, hardly would anyone believe that players in the Go game, as well as team games including most sport and online games, manage to find and employ a Nash Equilibrium strategy.

To take sub-optimality in experts' demonstrations into account, we propose a completely different approach to perform multi-agent IRL in zero-sum discounted stochastic games. We still assume experts should be performing decently well. Therefore, the margin between experts' performances and those of optimal strategies should be minimized by the reward function we yield, even though the performance gap is most likely above zero. Specifically, we take a game-theoretical perspective on competitive MDPs, and the proposed IRL algorithm alternates between two major steps. In the policy step, we find a Nash Equilibrium strategy for the game parametrized by the current reward function, while in the reward step we compare experts' performances against those of the Nash Equilibrium strategy, and update the reward function to minimize the performance gap. Our framework significantly departs from previous works since the model and algorithm do not decouple the agents. 

In the policy step, we propose an adversarial training algorithm to solve for Nash Equilibrium in zero-sum discounted stochastic games. We show the theoretical appeal of guarantee on global convergence to an optimal solution under this adversarial training formulation, as well as an algorithm based on deep neural networks for policy models. During the adversarial training we pit the agents against their optimal opponents, and perform actor-critic style Proximal Policy Optimization in \cite{schulman2017proximal} to improve the agents' policies. In the reward step, we use a deep neural network as well to model the reward function. We sample expert actions from demonstrations and optimal actions from models trained in the policy step, and then conduct stochastic gradient descent to minimize the performance gap between optimal strategies and experts. It is worth noting that other zero-sum IRL methodologies decouple the agents and solve two sub-problems (or impose two sets of independent constraints) due to their optimality assumption, while our algorithm always considers and relies on both agents in the game to solve for the reward function. 

Our major contributions are as follows. First, to the best of our knowledge, our IRL algorithm is drastically different from all previous competitive multi-agent IRL algorithms, as it is able to yield the reward function in zero-sum discounted stochastic games after abandoning the optimality assumption of expert demonstrations and it does not decouple the agents. Second, we propose an adversarial training algorithm to find a Nash Equilibrium policy in zero-sum stochastic games, and show the non-existence of local maxima in its objective function. Lastly, by utilizing deep neural networks and stating the algorithm accordingly, our multi-agent IRL approach addresses larger-scale games or tasks that are not amenable to previous methods, which rely on tabular representation and linear or quadratic programming. Our numerical experiments show that, compared with existing competitive multi-agent IRL methodologies, our approach manages to solve a large-scale problem and recover both the reward and policy functions robustly regardless of variation in the quality of expert demonstrations.

The rest of the paper is organized as follows. The literature review is followed by Section 3, where we present notation and basic results for zero-sum discounted games in the MDP formalism. In Section 4 we discuss our IRL algorithm, and in Section 5 we present the Nash Equilibrium algorithm in zero-sum discounted games and its theoretical analysis. Section 6 demonstrates performance of our algorithms by means of numerical experiments, and Section 7 discusses some directions for future work.

\section{Related Work}

In the area of reinforcement learning, an agent learns to improve its performance in a task by interacting with the environment and learning from its experiences. Usually, performance is defined by the cumulative reward the agent has obtained throughout the task. This suggests the necessity of knowledge about the reward function or the observed immediate reward. However, in some cases the reward function simply remains unknown or is implicit, and all that is accessible to the agents is a set of demonstrations about how some experts have performed in the task. Under such scenarios, one of the most straightforward approaches is to perform imitation learning in a supervised fashion. One such implementation is the DAgger algorithm \cite{ross2011reduction}. The downsides of the supervised imitation learning approach are: (1) continuous human guidance or interventions are needed sometimes (as in DAgger); (2) by formulating the task as essentially a regression problem and setting the objective as minimizing the loss of the prediction error, the agent might fail to handle a complex case, for example, experts employ a stochastic policy that cannot be approximated by a deterministic function.

On the contrary, IRL approaches aim at recovering the unknown reward function. As an inverse problem, IRL is ill-defined in most cases because it is not known beforehand how well the experts have performed in the task. Even if we assume that experts have employed optimal strategies, there might be multiple reward functions that could explain the demonstrations ideally (note that there always exist trivial solutions including reward $R \equiv 0$ for any state or action). Therefore, existing IRL algorithms formulate distinct objectives or impose different constraints. As discussed in \cite{choi2016density}, one dichotomy for classifying IRL methodologies is to consider the algorithm as a margin-based approach or a probabilistic one. A margin-based method solves IRL by maximizing the margin between experts' performances and those of other existing policies (\cite{abbeel2004apprenticeship} for instance), whereas a probabilistic method revolves around a probability distribution of possible reward functions, either by maximizing the probability of a reward function based on observed expert demonstrations (\cite{ziebart2008maximum} for instance), or matching the probability distribution of state-action frequency in experts' demonstrations as in \cite{choi2016density}. 

Due to the formulation of the problem, many IRL methodologies approach the task as an imitation learning problem in essence. This is especially the case if we either implicitly or explicitly assume experts' strategies to be optimal. Abbeel \& Ng \cite{abbeel2004apprenticeship} aim to find policies that are comparable to the best possible ones under any possible reward function, and thus translate IRL to a distribution matching problem using linear combinations of pre-set basis functions. The guided cost learning algorithm in \cite{finn2016guided}, initially thought of as an extension of the maximum entropy IRL method in \cite{ziebart2008maximum}, has been shown to bear a close connection with generative adversarial networks \cite{goodfellow2014generative} with the reward function being the discriminator and the sampling distribution (policy) as the generator. This connection partially explains why sometimes the recovery of the reward function fails, but the algorithm still obtains a valid policy function (as Finn et al. \cite{finn2016guided} observe in their experiments). Generative adversarial imitation learning \cite{ho2016generative}, as the name suggests, conducts imitation learning, and maintains a reward function for the sole purpose of providing a gradient for policies functions (similar to discriminators in generative adversarial networks).

All aforementioned works are in the single-agent setting. Our interest is in competitive multi-agent systems. To the best of our knowledge, existing IRL works concerning multi-agent competitive games make the same assumption that expert demonstrations are optimal. IRL can thus be decoupled into two sub-problems in two-agent games. Reddy et al. \cite{reddy2012inverse} base the whole algorithm on the optimality assumption on experts' strategies, and decouple a IRL problem in a general-sum game when updating the reward function. Lin et al. \cite{lin2017multi} apply the Bayesian IRL framework in \cite{ramachandran2007bayesian} to multi-agent zero-sum stochastic games, and assume that two agents in the game employ a unique minimax bi-policy under an unseen reward function. This Bayesian approach solves for the reward function only, and optimality for both agent's demonstrations leads to two sets of constraints. As a result of the distinct formulation, our model does not decouple the two agents. Instead, in our algorithm we always let the two agents compete against each other, and compare optimal performance against demonstrations to find the reward function.

As a subroutine in our IRL algorithm, measurement of optimal strategies' performances entails the knowledge about Nash Equilibrium strategies as we explicitly compare experts' demonstrations against optimal strategies. On solving for Nash Equilibrium in a stochastic game, the past decade has seen new algorithms including \cite{akchurina2009multiagent}, \cite{prasad2015two}, and \cite{prasad2015study}. Unfortunately, most of these methods do not suit our case. For instance, the optimization-based ones rely on enumeration of state-action pairs of the game, which is infeasible for large games. Algorithm in \cite{prasad2015two} solves for Nash Equilibrium in general-sum stochastic games, but in the zero-sum case the agents always receive a 0 gradient because the update steps rely on the sum of returns for both agents, which is fixed at 0 in zero-sum games. In order to efficiently handle zero-sum games we thereby propose and use an adversarial training algorithm that finds a Nash Equilibrium strategy using deep neural networks as model approximations.

\section{Preliminaries}

\subsection{Zero-Sum Discounted Stochastic Game}

From the field of competitive MDPs, we adopt the zero-sum discounted stochastic game setting as the underlying framework for the current study. Formally, a zero-sum stochastic game is defined by a tuple $\langle \mathcal{S},\mathcal{A}^f,\mathcal{A}^g,R,P,\gamma \rangle$, where the quantities are defined as follows.

\begin{itemize}
\item \textbf{Agents and Strategies:} $f,g$ denote strategies used by the two agents in the game. Note that the first order Markovian Property holds for all the strategies involved in competitive MDPs mentioned in this study. Details on strategy functions are given in Section 3.2. We slightly abuse the notation and also use $f,g$ to indicate the two agents in the game.
\item \textbf{States:} $\mathcal{S}$ is a finite set of $N_{\mathcal{S}}$ distinct states (each being a real vector) that can be visited in the game.
\item \textbf{Actions:} $\mathcal{A}^i = \{\mathcal{A}^i(s)  |s \in \mathcal{S}\}, i = f,g$ defines the action space for agent $f, g$. For each state $s \in \mathcal{S}$,  $\mathcal{A}^i(s)$ is a finite, discrete set containing all the available actions (each being a real vector) for agent $i$ at state $s$. If the set of candidate actions is the same at every state (which we assume for ease of exposition), then we write $\mathcal{A}^i = \{a^i(1), a^i(2), ..., a^i(|\mathcal{A}^i|)\}, i = f,g$ to refer to the fixed available action set for agent $i$.
\item \textbf{Rewards:} $R: \mathcal{S} \rightarrow \mathbb{R} $ is reward function for the game. In accordance with the zero-sum nature of the game, $R(s)$ denotes reward received by agent $f$ while $-R(s)$ for agent $g$.
\item \textbf{States Transition Probabilities:} $P: \mathcal{S} \times \mathcal{A}^f \times \mathcal{A}^g \times \mathcal{S} \rightarrow \mathbb{R}$  is the state transition probability function of the game. If agents $f,g$ take actions $a^f \in \mathcal{A}^f, a^g \in \mathcal{A}^g$ respectively at state $s \in \mathcal{S}$, then $s^\prime \in \mathcal{S}$ is the next visited state with probability $P(s^\prime|s,a^f,a^g)$. Note that in an MDP, the transition probability depends only on the current state and agents' actions.
\item \textbf{Discount Factor:} $\gamma \in [0,1)$ is the discount factor on cumulative reward.
\end{itemize}

\subsection{Strategies and State Value Functions }

In a competitive MDP, the two agents $f,g$ employ Markovian strategies $f(a^f|s),\text{ }g(a^g|s)$ respectively to interact with the environment and compete against each other. The Markovian property of strategies states that agents choose actions based only on $s$, the current state of the game, and all the other historical information is disregarded. Following the definition of a probability distribution, $f(a^f|s),\text{ }g(a^g|s)$ should be non-negative and sum up to 1 over the action space. A pure strategy means that, for any state $s$, agents choose only one action deterministically, and the probability distribution $f(a^f|s)$ (or $g(a^g|s)$ ) is fixed as 0 for all other candidate actions. On the contrary, a mixed strategy means that distribution $f$ or $g$ can assign a non-zero probability to more than one candidate actions.

The state value function measures the discounted cumulative return. Given strategy functions $f\text{ and }g$, the state value function (for $f$) at state $s_0$ is the expected discounted return
\begin{align}
v^{f,g}(s_0;R) = \mathbb{E}_{ a^f_t \sim f(a^f|s_t),\text{ } a^g_t \sim g(a^g|s_t),\text{ } s_{t+1} \sim P(s^\prime|s_t,a^f_t,a^g_t) }\Big[ \sum_{t=0}^{\infty} \gamma^t R(s_t) \Big],\label{v}
\end{align}
while the expected discounted cumulative return for $g$ is $-v^{f,g}(s_0;R)$. Since we consider IRL, we show dependency on $R$ explicitly as $R$ changes in our IRL algorithm.

\subsection{Q-Function and Advantage Function for Actions}

Aside from the state value function $v^{f,g}(s;R)$, another useful metric is the state-action value function (namely, Q-function)
\begin{align}
Q^{f,g}_f(s,a;R) = R(s) + \gamma \mathbb{E}_{ a^g \sim g(a|s), \text{ } s^\prime \sim P(s^\prime|s,a,a^g)} \Big[ v^{f,g}(s^\prime;R) \Big], \label{qf} \\
Q^{f,g}_g(s,a;R) = -R(s) + \gamma \mathbb{E}_{a^f \sim f(a|s),  \text{ } s^\prime \sim P(s^\prime|s,a^f,a)} \Big[ -v^{f,g}(s^\prime;R) \Big]. \label{qg}
\end{align}

Based on the state value function and Q-function, the advantage function measuring benefits of actions over current strategies is defined as
\begin{align}
A^{f,g}_f(s,a^f;R) = Q^{f,g}_f(s,a^f;R) - v^{f,g}(s;R), \label{af} \\
A^{f,g}_g(s,a^g;R) = Q^{f,g}_g(s,a^g;R) + v^{f,g}(s;R). \label{ag} 
\end{align}

% Q-Function and advantage function mentioned above are generalized from the uni-agent case, where one and only one agent acts under Markov strategy $\pi(a|s)$ in a stationary environment with state transition probability $P(s^\prime|s,a)$. In uni-agent case, $v$, $Q$, and $A$ functions are defined as
% $$v^{\pi}(s_0;R) = \mathbb{E}_{a_t\sim\pi(a|s_t),\text{ }s_{t+1}\sim P(s^\prime|s_t,a_t)} \Big[\sum_{t=0}^{\infty}\gamma^t R(s_t) \Big],$$
% $$Q^{\pi}(s,a;R) = R(s) + \gamma \mathbb{E}_{a\sim\pi(a|s),\text{ } s^\prime \sim P(s^\prime|s,a)}v^{\pi}(s^\prime), $$
% $$A^{\pi}(s,a; R) = Q^{\pi}(s,a;R) - v^{\pi}(s;R)  .$$

The advantage functions can be used for characterization of optimal strategies in both single-agent and multi-agent MDPs.

% , as well as the concept of optimality in Markov decision processes: advantage functions can be used to characterize optimal strategies. Consider for a uni-agent game, there exists an optimal strategy $\pi^*$ that maximizes state value for each state. Connection between state value $v^{\pi^*}$ and state-action value $Q^{\pi^*}$ is explained by Bellman's optimality equation
% $$v^{\pi^*}(s;R) = \max_{a \in \mathcal{A}} Q^{\pi^*}(s,a;R) \text{ for } \forall s \in \mathcal{S}, $$
% which implies that advantage should be zero for selected actions, and non-positive for unselected actions. Therefore, one can use advantage function to characterize optimal strategies in a stationary environment (uni-agent MDP).

\subsection{Nash Equilibrium in a Zero-Sum Discounted Stochastic Game}

To proceed with the discussion, we need a formal definition on optimal strategies in a zero-sum stochastic game. In both zero-sum and general-sum two-person discounted stochastic games, when agents are allowed to play mixed Markovian strategies, it can be shown that there exists at least one pair of strategies $(f^*(R),g^*(R))$ that is optimal as $(f^*(R),g^*(R))$ reaches Nash Equilibrium under reward function $R$ (sometimes the term ``optimal strategy'' is used instead in the case of zero-sum discounted stochastic games). In the zero-sum case, a Nash Equilibrium strategy pair $(f^*(R),g^*(R))$ ensures that, for any $s \in S$ and any strategy pair $(f,g)$ we have
\begin{align}
v^{f^*(R),g}(s;R) \geq v^{f^*(R),g^*(R)}(s;R) \geq v^{f,g^*(R)}(s;R). \label{nasheq_ineq}
\end{align}

Therefore, $f^*(R)$ is a solution to optimization problem 
\begin{align}
\max_{f} \min_{g} \frac{1}{N_s} \sum_{s \in S}v^{f,g}(s;R), \label{obj_f}
\end{align}
and $g^*(R)$ is a solution to
\begin{align}
\min_{g} \max_{f} \frac{1}{N_s} \sum_{s \in S}v^{f,g}(s;R). \label{obj_g}
\end{align}
% This is due to the fact that, in (2), Nash Equilibrium $(f^*(R),g^*(R))$ forms a saddle point for any state $s$. 

As already mentioned above, we can characterize Nash Equilibriums in zero-sum discounted games using advantage functions. To be specific, $\big(f^*(R),g^*(R)\big)$ is a Nash Equilibrium if and only if for any $s$,
$$ A^{f^*(R),g^*(R)}_f(s,a; R) \leq 0 \text{ for } \forall a \in \mathcal{A}^f, \text{ and } A^{f^*(R),g^*(R)}_f(s,a; R) = 0 \text{ if }  f^*(R)(a|s)>0;$$
$$ A^{f^*(R),g^*(R)}_g(s,a; R) \leq 0 \text{ for } \forall a \in \mathcal{A}^g, \text{ and } A^{f^*(R),g^*(R)}_g(s,a; R) = 0 \text{ if }  g^*(R)(a|s)>0.$$

% In the perspective of advantage functions, at Nash Equilibrium, every action has a non-positive advantage, namely $A^{f^*(R),g^*(R)}_{f}(s,a^f;R) \leq 0, A^{f^*(R),g^*(R)}_{g}(s,a^g;R) \leq 0$, suggesting that Bellman's optimal equation holds and both agents cannot further improve their discounted cumulative returns.

Uniqueness of Nash Equilibrium is not guaranteed. Therefore, technically speaking $f^*(R)$ and $g^*(R)$ are multivalued functions, mapping to a set that contains all Nash Equilibrium strategies. In \eqref{nasheq_ineq} the strategy appearing at superscripts is just one candidate from the entire set. Fortunately, this does not pose extra issues since the value of the game 
$$v^*(s) = \max_{f} \min_{g} v^{f,g}(s;R)$$
is known to be unique for any $s \in \mathcal{S}$ in a zero-sum discounted stochastic game. Videlicet, in zero-sum cases all Nash Equilibrium strategy pairs lead to the same state value functions, and \eqref{nasheq_ineq} always holds.

%Objective function (3) inspires us that solving Nash Equilibrium strategies in a zero-sum stochastic game can be treated as an adversarial training problem. The outline of the algorithm is as follows: in order to obtain $f^*$, structure of the game (reward, state transition probabilities, and action sets) is required as input; the algorithm alternates between occasional training of an agent ($f$ step) and the frequent update of its "best possible opponent" ($g$ step); if $f$ step can not further improve the agent, then the algorithm has converged and $f^*$ has been reached as the output of the algorithm. The algorithm is detailed in next section. 

%one may wonder whether it is feasible to train two models $f$ and $g$ simultaneously, and lead $f$ to converge to $f^*$ and $v^{f,g}$ to $v^*$. In the following sections, we'll first propose an adversarial training algorithm aiming to solve the optimal strategies and value of the game, and then we'll show that for objective function (3) there exists no local maximum for $f$, and convergence to optimal strategy $f^*$ is ensured when training is conducted ideally.

\section{Competitive Multi-agent Inverse Reinforcement Learning with Sub-Optimal Demonstrations}

\subsection{Inverse Reinforcement Learning Problem}

The purpose of our work is to address inverse reinforcement learning (IRL) problems in zero-sum stochastic games. As defined by Russel in \cite{russell1998learning}, an IRL algorithm relies on the knowledge of
\begin{itemize}
\item a model of the environment, i.e., state transition function $P(s^\prime|s,a^f,a^g)$ is known; and
\item demonstrations of agents' behaviors and visited states. An observation set 
$$\mathcal{D} = \{(s_i, a^{E,f}_i, a^{E,g}_i)\text{ }|\text{ }i=1,2,3,....,N_{\mathcal{D}}\}$$
is available, and shows how some experts have performed in the task. Here $s_i$ is the visited state, and $a^{E,f}_i, a^{E,g}_i$ stand for actions taken by the two expert players where we use $E$ to denote experts. 
\end{itemize}

Note that in $\mathcal{D}$, the next visited state is not required since we already have access to state transition probabilities. Besides, subscript $i$ does not necessarily stand for time, and each observation $(s_i, a^{E,f}_i, a^{E,g}_i)$ can be drawn from different rounds of games so that the $i$-th observation can be irrelevant to the $(i-1)$-th or $(i+1)$-th observation. Lastly,  we do not attempt to model the expert strategies that have generated $a^{E,f}_i, a^{E,g}_i$. Instead, we only sample observations in $\mathcal{D}$ and compare their performance with that of optimal strategies.

\subsection{Objective Function}

The formulation of an IRL problem hinges on the choice of the objective function. As stressed before, we do not assume the optimality of experts' demonstrations, but we still assume experts to have demonstrated performances that are comparable with the best possible strategies in this game. More specifically, based on inequality \eqref{nasheq_ineq} we have
\[ v^{f^*(R),g^{E}|_{\mathcal{D}}}(s;R) \geq v^{f^*(R),g^*(R)}(s;R) \geq v^{f^{E}|_{\mathcal{D}},g^*(R)}(s;R)\text{ for }\forall s \in \mathcal{S}. \]
Here $R$ is the reward function we aim to solve for, and $f^{E}|_{\mathcal{D}}$ is a policy that concurs with actions $a^{E,f}_i$ for each observed state $s_i,\text{ }i=1,2,\ldots,N_{\mathcal{D}}$. More formally, 
$$f^{E}|_{\mathcal{D}}(a|s) = \dfrac{\#\{(s,a,\_)\}}{\#\{(s,\_,\_)\}} \text{ if }\#\{(s,\_,\_)\}>0$$ where $\#$ denotes the count of occurrence in $\mathcal{D}$. For states $s$ with $(s,\_,\_) \not\in \mathcal{D}$, we remain agnostic about $f^{E}|_{\mathcal{D}}$. We similarly define $g^{E}|_{\mathcal{D}}(a|s) = \dfrac{\#\{(s,\_,a)\}}{\#\{(s,\_,\_)\}}$ when the denominator is positive.
% \[
%     f^{E}|_{\mathcal{D}}(a|s_i)= 
% \begin{cases}
%     \dfrac{\#\{(s,a,\_)\}}{\#\{(s,\_,\_)\}} & \text{if } (s_i,a,\_)\in \mathcal{D}\\
%     0              & \text{otherwise}.
% \end{cases}
% \]

Meanwhile, we would like margins
\begin{align}
\mathbb{E}_s\Big[ v^{f^*(R),g^{E}|_{\mathcal{D}}}(s;R) - v^{f^*(R),g^*(R)}(s;R)\Big], \label{obj_irl_p1} \\
\mathbb{E}_s\Big[ v^{f^*(R),g^*(R)}(s;R) - v^{f^{E}|_{\mathcal{D}},g^*(R)}(s;R)\Big] \label{obj_irl_p2}
\end{align}
to be reasonably tight. By summing up the two margins, we yield the optimization problem for our IRL model
\begin{align}
\min_{R}\min_{f^{E}|_{\mathcal{D}}, g^{E}|_{\mathcal{D}}}  \mathbb{E}_{s}\Big[ v^{f^*(R),g^{E}|_{\mathcal{D}}}(s;R) - v^{f^{E}|_{\mathcal{D}},g^*(R)}(s;R)\Big] \label{obj_IRL}
\end{align}
$$\text{where } f^*(R) \in \operatorname*{argmax}_{f} \min_{g} \frac{1}{N_s} \sum_{s \in S}v^{f,g}(s;R), $$
$$ \text{and }g^*(R) \in \operatorname*{argmin}_{g} \max_{f} \frac{1}{N_s} \sum_{s \in S}v^{f,g}(s;R). $$

Note that the minimization on $f^{E}|_{\mathcal{D}}, g^{E}|_{\mathcal{D}}$ in \eqref{obj_IRL} is required since $f^{E}|_{\mathcal{D}}, g^{E}|_{\mathcal{D}}$ are uniquely defined only on $\mathcal{D}$. The model encourages $f^{E}|_{\mathcal{D}}$ and $g^{E}|_{\mathcal{D}}$ outside of $\mathcal{D}$ to follow the actions in $f^*(R)$ and $g^*(R)$.

\subsection{Algorithm}

We approach \eqref{obj_IRL} by an iterative algorithm. In each iteration,  we find Nash Equilibrium strategy $f^*(R), g^*(R)$ given current reward function $R$, then $R$ is updated based on incumbent $f^*(R), g^*(R)$, which requires the estimation of (9) and (10). Regarding the minimization operation on $f^{E}|_{\mathcal{D}}, g^{E}|_{\mathcal{D}}$ in \eqref{obj_IRL}, on states that are not demonstrated in by $\mathcal{D}$ the algorithm considers only strategies equal to the current $f^*(R), g^*(R)$. Furthermore, we consider only $f^{E}|_{\mathcal{D}}, g^{E}|_{\mathcal{D}}$ that match with $\mathcal{D}$ on a sample random demonstration, and follow the current $f^*(R), g^*(R)$ otherwise, respectively. In doing so we avoid the need to deal with $\min_{f^{E}|_{\mathcal{D}}, g^{E}|_{\mathcal{D}}}$, and essentially the algorithm updates only the reward function $R$ and Nash Equilibrium.

% Two problems remain unsolved before implementing any algorithm to solve objective function (4): solving $f^*(R), g^*(R)$ given $R$, and the estimation of $\mathbb{E}_s v^{f^*(R),E}(s;R)$, $ \mathbb{E}_s v^{f^*(R),g^*(R)}(s;R)$ using observation set $\mathcal{D}$. As for the former problem, we propose an adversarial training algorithm to solve Nash Equilibrium strategies, and our IRL algorithm would alternate between Nash Equilibrium step where $R$ is held fixed, and the reward function step during which the current approximation of $f^*(R), g^*(R)$ would be used to find a better $R$ that minimizes performance gaps in (4). Details on Nash Equilibrium step will be covered in section 4 and algorithm 2.

As for solving Nash Equilibrium in zero-sum discounted games, the proposed adversarial training algorithm is detailed in Algorithm 2 and Section 4. Next we focus on the update step on the reward function with current $f^*(R), g^*(R)$. 

Note that $(f^{E}|_{\mathcal{D}}, g^{E}|_{\mathcal{D}})$ poses inconveniences since the expert policy is unknown when a state $s$ outside of $\mathcal{D}$ is visited. Our workaround is to let experts act only at the very first step, and use $(f^*,g^*)$ for the following states. Such a treatment follows the logic that, by bounding the performance gap at the first step (equivalent to bounding advantage of actions), we also bound the gap of discounted cumulative reward for the infinitely many steps. 

Specifically, $v^{f^{E}|_{\mathcal{D}},g^*(R)}(s;R)$ is estimated by sampling trajectories $(s^{E,g^*(R)}_1,\ldots,s^{E,g^*(R)}_T)$ of a fixed length $T$ in the following manner (note that $E$ stands for experts):
\begin{equation} \label{sample_1and2}
\begin{gathered}
(s^{E,g^*(R)}_0,a^f_0,\_) \sim \mathcal{D}, \\ 
a^g_t \sim g^*(R)(a|s^{E,g^*(R)}_t)\text{ for }0\leq t \leq T, \\
a^f_t \sim f^*(R)(a|s^{E,g^*(R)}_t)\text{ for }1 \leq t \leq T, \\
\text{ and } s^{E,g^*(R)}_t \sim P(s^\prime|s^{E,g^*(R)}_{t-1},a^{f}_{t-1},a^{g}_{t-1})\text{ for }1 \leq t \leq T.
\end{gathered}
\end{equation}

For the estimation of $v^{f^*(R),g^{E}|_{\mathcal{D}}}(s;R)$, trajectories $(s^{f^*(R),E}_1,s^{f^*(R),E}_2,\ldots,s^{f^*(R),E}_T)$ are sampled in a symmetric fashion, where the initial state $s^{f^*(R),E}_0$ and initial action for $g$ $a^g_0$ is sampled from $\mathcal{D}$, while the other actions and states are generated by $f^*(R), g^*(R)$. 

The algorithm is shown in Algorithm 1. We use deep neural networks as model approximations for policies and reward functions involved in the IRL task. To model the reward function, a deep neural network $R_{\theta_R}(s)$ is used and parametrized by $\theta_R$. Similarly, two deep neural nets $f_{\theta_f}(a|s)$, $g_{\theta_g}(a|s)$ are maintained and updated in the algorithm to approximate $f^*(R_{\theta_R})$, $g^*(R_{\theta_R})$. A regularization term $\phi(\theta_R)$ is added to prevent trivial solutions ($R(s) \equiv 0$ for instance) and normalize the scale of $R(s)$. 

As shown in Algorithm 1, in the policy step at steps 3 and 4, we update $\theta_f, \theta_g$ under the current $R_{\theta_R}$ and try to find a Nash Equilibrium. Every $K_R$ iterations, we check the performance of Nash Equilibrium policies in step 6. (The estimation of $\hat{v}^{f^{\text{best},g}},\hat{v}^{f,g^{\text{best}}}$ is detailed in Section \ref{NashEq_section}.) To ensure a good approximation of optimal policies under the incumbent reward function, the policy step is performed much more frequently than the reward step, and we update $\theta_R$ only when we believe $f_{\theta_f}$ and $g_{\theta_g}$ approximate $f^*(R_{\theta_R}), g^*(R_{\theta_R})$ well enough. When the current $f_{\theta_f}, g_{\theta_g}$ are accurate enough compared against a threshold $\tau$, we perform $I_{R}$ times a reward step that minimizes the performance gap between $f^*(R),g^{E}|_{\mathcal{D}}$ and $f^{E}|_{\mathcal{D}},g^*(R)$ with regularization term $\phi(\theta_R)$ based on trajectories generated as described in \eqref{sample_1and2}. Note that the policy step is performed at each iteration, and the only difference between each policy step is that different trajectories are sampled and used to update the Nash Equilibrium policy under current $R_{\theta_R}$.

\begin{algorithm}[ht]
%   \algsetup{linenosize=\tiny}
%    \scriptsize
  \caption{Inverse Reinforcement Learning in Zero-Sum Discounted Stochastic Games}
  \begin{algorithmic}[1]
    \Require
      \Statex Observed experts demonstrations $\mathcal{D} = \{(s_i,a^f_i,a^g_i)\text{ }|\text{ }i = 1,2,\ldots,N_{\mathcal{D}}\}$;
      \Statex Positive integers $K_{R},I_{R}$; Nash Equilibrium threshold $\tau$; learning rate $\lambda$.
    \Statex
\Statex \textbf{Initialize:} Parameters $\theta_R$ for the reward function, and $\theta_f,\text{ }\theta_g$ to parametrize $f_{\theta_f}(a|s),\text{ }g_{\theta_g}(a|s)$ for Nash Equilibrium policies

\For{$\text{i} = 1, 2, 3, \ldots $} \Comment{policy step}

\State Update $\theta_f$ to find Nash Equilibrium strategy for $f$ under current $R_{\theta_R}$, return also $\hat{v}^{f,g^{\text{best}}}$ 
\State Update $\theta_g$ to find Nash Equilibrium strategy for $g$ under current $R_{\theta_R}$, return also $\hat{v}^{f^{\text{best}},g}$ 

\If{$i\text{ } \% \text{ }K_{R} = 0 $}

\If{$\hat{v}^{f^{\text{best}},g} - \hat{v}^{f,g^{\text{best}}} < \tau$ } \Comment{check performances of Nash Equilibrium policies}

\For{$\text{j} = 1,2,3,\ldots,I_R$} \Comment{reward step}
\State Sample one observation from $\mathcal{D}$: $(s,a^{E,f},a^{E,g})$
\State Use \eqref{sample_1and2} to obtain $\{s^{f^*(R_{\theta_R}),E}_1,s^{f^*(R_{\theta_R}),E}_2,\ldots,s^{f^*(R_{\theta_R}),E}_T\}, \{s^{E,g^*(R_{\theta_R})}_1,s^{E,g^*(R_{\theta_R})}_2,\ldots,s^{E,g^*(R_{\theta_R})}_T\}$

\State $\hat{v}^{f}(\bar{\theta_R}) \leftarrow R_{\bar{\theta_R}}(s) + \sum_{t=1}^T \gamma^{t-1} R_{\bar{\theta_R}}\Big(s^{f^*(R_{\theta_R}),E}_t\Big)$
\State $\hat{v}^{g}(\bar{\theta_R}) \leftarrow R_{\bar{\theta_R}}(s) + \sum_{t=1}^T \gamma^{t-1} R_{\bar{\theta_R}}\Big(s^{E,g^*(R_{\theta_R})}_t\Big)$
\State $\theta_R \leftarrow \theta_R - \lambda\nabla_{\bar{\theta_R}}\Big( \hat{v}^{f}(\bar{\theta_R}) - \hat{v}^{g}(\bar{\theta_R}) + \phi(\bar{\theta_R}) \Big)|_{\bar{\theta_R} = \theta_R}$
\EndFor

\EndIf

% \State Sample one observation from $\mathcal{D}$: $(s,a^{E,f},a^{E,g})$
% \State Use \eqref{sample_1} to obtain a sequence $\{s^{f^*(R_{\theta_R}),E}_1,s^{f^*(R_{\theta_R}),E}_2,\ldots,s^{f^*(R_{\theta_R}),E}_T\}$ 
% \State Use \eqref{sample_2} to obtain a sequence $\{s^{E,g^*(R_{\theta_R})}_1,s^{E,g^*(R_{\theta_R})}_2,\ldots,s^{E,g^*(R_{\theta_R})}_T\}$

% \State $\hat{v}^{f}(\bar{\theta_R}) \leftarrow R_{\bar{\theta_R}}(s) + \sum_{t=1}^T \gamma^{t-1} R_{\bar{\theta_R}}\Big(s^{f^*(R_{\theta_R}),E}_t\Big)$
% \State $\hat{v}^{g}(\bar{\theta_R}) \leftarrow R_{\bar{\theta_R}}(s) + \sum_{t=1}^T \gamma^{t-1} R_{\bar{\theta_R}}\Big(s^{E,g^*(R_{\theta_R})}_t\Big)$
% \State $\theta_R \leftarrow \theta_R - \lambda\nabla_{\bar{\theta_R}}\Big( \hat{v}^{f}(\bar{\theta_R}) - \hat{v}^{g}(\bar{\theta_R}) + \phi(\bar{\theta_R}) \Big)|_{\bar{\theta_R} = \theta_R}$
\EndIf

\EndFor
    
  \end{algorithmic}
\end{algorithm}

\section{Solving Nash Equilibrium in Zero-Sum Discounted Stochastic Games}\label{NashEq_section}

\subsection{Algorithm}

In order to find Nash Equilibrium in zero-sum stochastic games, we adopt the framework of generative adversarial network in \cite{goodfellow2014generative}, and propose an algorithm that is used in the policy step in Algorithm 1 at steps 3 and 4. The algorithm finds a Nash equilibrium strategy for only one agent. We present the algorithm for solving $f^*(R)$, and $g^*(R)$ can be solved in a similar fashion. In the remainder of this section, we omit the dependency on $R$, which is fixed in the policy step. 

The definition of Nash Equilibrium in \eqref{nasheq_ineq} in zero-sum discounted games naturally suggests adversarial training as a solution methodology. Based on \eqref{nasheq_ineq},
we first identify the best response $g$ to current $f$, and then update $f$ marginally to compete against the best response $g$. We repeat these two steps until $f^*$ has been reached.

Two deep neural networks $f_{\theta_f}(s)$ and $g_{\theta_g}(s)$ are used, parametrized by $\theta_f$ and $\theta_g$ respectively. Both networks take state vector $s \in \mathcal{S}$ as input, which is completely public to both sides. Each network outputs a probability distribution over action space, namely
$$f_{\theta_f}(s) = \Big(\text{ }f_{\theta_f}(a^f(1)\text{ }|\text{ }s),\text{ }f_{\theta_f}(a^f(2)\text{ }|\text{ }s),\text{ }\ldots,\text{ }f_{\theta_f}(a^f({|\mathcal{A}^f|})\text{ }|\text{ }s)\text{ }\Big),$$
$$g_{\theta_g}(s) = \Big(\text{ }g_{\theta_g}(a^g(1)\text{ }|\text{ }s),\text{ }g_{\theta_g}(a^g(2)\text{ }|\text{ }s),\ldots,\text{ }\text{ }g_{\theta_g}(a^g({|\mathcal{A}^g|})\text{ }|\text{ }s)\text{ }\Big).$$
Agents then sample actions from the probability distribution and act accordingly. 

As for policy gradient methods used in our algorithm, we choose the actor-critic style Proximal Policy Optimization algorithm (PPO) from \cite{schulman2017proximal} because of its superior and stable performances. To perform actor-critic style PPO, a state value model is required and we denote it as $v^{f,g}_{\theta_v}(s)$. Again, $v^{f,g}_{\theta_v}(s)$ is a deep neural network that takes state vector $s$ as input, and outputs a scalar as the estimation for state value defined in (1).

The online training on $f$ and $g$ relies on $T$-step trajectories of the game. From a randomly initialized $s_0$, we run the agents for $T$ consecutive steps to obtain a trajectory as an ordered tuple 
$$\Big( s_0, s_1, \ldots, s_{T-1}, s_{T} \Big). $$
At each state $s_t$, $a^f_t,\text{ }a^g_t$ are the actions taken by $f,g$ based on the incumbent policy parameters, $R_t$ is the immediate reward received by $f$ at step $t$, and $s_{t+1}$ is the next state visited by agents after their actions.

Similar to \cite{schulman2017proximal}, a set of target networks is maintained for generating trajectories, while the training step based on observed trajectories updates the original networks. Parameters for target networks are denoted as $(\theta^{\text{target}}_f,\theta^{\text{target}}_g,\theta^{\text{target}}_{v,f},\theta^{\text{target}}_{v,g})$. After every $K_{\text{refresh}}$ iterations, they are periodically refreshed by $(\theta_f,\theta_g,\theta_{v,f},\theta_{v,g})$ we are training. Estimation of advantage for $f$ at each step is
\begin{equation} \label{adv_PPO}
\begin{gathered}
\hat{A}^f_t = \delta^f_t + (\gamma \lambda) \delta^f_{t+1} + ... + (\gamma \lambda)^{T-t-1} \delta^f_{T-1} \text{ }\text{ for }\text{ } t = 0,1,2,...,T-1, 
\nonumber \\
 \delta^f_t = R_t + \gamma v^{f,g}_{\theta^{\text{target}}_{v,f}}(s_{t+1}) -  v^{f,g}_{\theta^{\text{target}}_{v,f}(s_t)}, \nonumber
\end{gathered}
\end{equation}

and the clipped loss in PPO for $f$ is
\begin{equation} \label{loss_f_PPO}
\begin{gathered}
L^{f,\text{CLIP}}(\theta_f) = -\sum_t \min\Big(r^f_t(\theta_f)\hat{A}^f_t,(1-\epsilon)\hat{A}^f_t,(1+\epsilon)\hat{A}^f_t\Big),  \nonumber
\\
r^f_t(\theta_f) = \frac{ f_{\theta_f}(a^f_t|s_t) }{f_{\theta^{\text{target}}_f}(a^f_t|s_t)}. \nonumber
\end{gathered}
\end{equation}

Similarly, we formulate the clipped loss for $g$ as
\begin{equation} \nonumber\label{adv_g_PPO}
\begin{gathered}
\hat{A}^g_t = \delta^g_t + (\gamma \lambda) \delta^g_{t+1} + ... + (\gamma \lambda)^{T-t-1} \delta^g_{T-1} \text{ }\text{ for }\text{ } t = 0,1,2,...,T-1, 
\\
 \delta^g_t = -R_t + \gamma v^{f,g}_{\theta^{\text{target}}_{v,g}}(s_{t+1}) -  v^{f,g}_{\theta^{\text{target}}_{v,g}}(s_t),
\end{gathered}
\end{equation}
\begin{equation} \nonumber \label{loss_g_PPO}
\begin{gathered}
L^{g,\text{CLIP}}(\theta_g) = -\sum_t\min\Big(r^g_t(\theta_f)\hat{A}^g_t,(1-\epsilon)\hat{A}^g_t,(1+\epsilon)\hat{A}^g_t\Big), 
\\
r^g_t(\theta_{g}) = \frac{ g_{\theta_g}(a^g_t|s_t) }{g_{\theta^{\text{target}}_g}(a^g_t|s_t)}.
\end{gathered}
\end{equation}

The loss for updating $\theta_{v,f},\theta_{v,g}$ is
\begin{equation} \nonumber \label{loss_v_PPO}
\begin{gathered}
L^v(\theta_{v,f},\theta_{v,g}) = \sum_t \Big[ ( v^{f,g}_{\theta_{v,f}}(s_t) - v^{f,\text{target}}_t )^2 + ( v^{f,g}_{\theta_{v,g}}(s_t) - v^{g,\text{target}}_t )^2  \Big], 
\\
v^{f,\text{target}}_t = v^{f,g}_{\theta^{\text{target}}_{v,f}}(s_t) + \hat{A}^f_t,
\text{ }v^{g,\text{target}}_t = v^{f,g}_{\theta^{\text{target}}_{v,g}}(s_t) + \hat{A}^g_t.
\end{gathered}
\end{equation}

Lastly, as requested in step 6 in Algorithm 1, periodically we need to estimate the performance of $f_{\theta_f}$ when competing against its best adversarial $g_{\theta_g}$. Each time, we sample a batch of 64 $T$-step trajectories $\mathcal{T}$, and calculate the average of discounted cumulative return on these trajectories to yield
\begin{equation} \label{v_f_g_best}
\begin{gathered}
\hat{v}^{f,g^{\text{best}}} = \frac{1}{|\mathcal{T}|}\sum_{ ( s_0, s_1, \ldots, s_{T-1}, s_{T} ) \in \mathcal{T} } \sum_{t=0}^T \gamma^t R(s_t).
\end{gathered}
\end{equation}

The full algorithm is shown in Algorithm 2. As we have mentioned, the proposed algorithm is similar to GAN since $g$ is updated at most of the iterations while in every $K_{\text{cycle}}$ iterations only a small proportion of them are meant for $f$ step. Though we do not show this in the algorithm, we recommend the use of a batch of agents running in parallel for collecting gradients (as in \cite{schulman2017proximal}). Lastly, we reiterate that Algorithm 2 is meant for finding $f^*(R)$. The training for $g^*$ is conducted separately in a similar fashion, and we estimate $\hat{v}_{f^{\text{best}},g}$ similarly as in \eqref{v_f_g_best}. 

\begin{algorithm}[ht]
  \caption{Adversarial Training Algorithm for Solving $f^*(R)$ in Zero-Sum Games}
  \begin{algorithmic}[1]
    \Require
      \Statex Integers $K_g,K_{\text{cycle}}, K_{\text{refresh}}$; learning rates $\lambda_g$, $\lambda_f$; horizon length $T$.
    \Statex
\State \textbf{Initialize: } Parameters $\theta_f,\theta_g$ for policy models and $\theta_v$ for state value model.
\State Set $\theta^{\text{target}}_f \leftarrow \theta_f$, $\theta^{\text{target}}_g \leftarrow \theta_g$, and $\theta^{\text{target}}_v \leftarrow \theta_v$ .

\For{$i = 1, 2, 3, .... $}

\State Randomly initialize the starting state $s_0$.
\State From initial state $s_0$, run $f_{\theta^{\text{target}}_f}(a|s),\text{ } g_{\theta^{\text{target}}_g}(a|s)$ for $T$ steps
\State Calculate estimated advantages for player $f$ and $g$: $\hat{A}^{f/g}_0,\hat{A}^{f/g}_1,...,\hat{A}^{f/g}_{T-1}$.

\If{$i\text{ } \% \text{ }K_{\text{cycle}} \leq K_g  $} \Comment{$g$ step}
\State $\theta_g \leftarrow \theta_g - \lambda_g \nabla_{\theta} L^{g,\text{CLIP}}(\theta)|_{\theta = \theta_g}$
\State $\theta_{v,f} \leftarrow \theta_{v,f} - \lambda_f \nabla_{\theta} L^v(\theta,\theta^\prime)|_{\theta = \theta_{v,f},\theta^\prime = \theta_{v,g}}$
\State $\theta_{v,g} \leftarrow \theta_{v,g} - \lambda_g \nabla_{\theta^\prime} L^v(\theta,\theta^\prime)|_{\theta = \theta_{v,f},\theta^\prime = \theta_{v,g}}$
\Else \Comment{$f$ step}
\State  $\theta_f \leftarrow \theta_f - \lambda_f \nabla_{\theta} L^{f,\text{CLIP}}(\theta)|_{\theta = \theta_f}$
\EndIf

\If{$i\text{ } \% \text{ }K_{\text{refresh}}\text{ } = 0  $} \Comment{refresh target networks}
\State Set $ \theta_f \leftarrow \theta^{\text{target}}_f$, $\theta_{g}  \leftarrow \theta^{\text{target}}_{g}$, $\theta_{v,f} \leftarrow  \theta^{\text{target}}_{v,f} $ , and $\theta_{v,g} \leftarrow  \theta^{\text{target}}_{v,g} $ .
\EndIf

\EndFor
  \end{algorithmic}
\end{algorithm}

% \begin{algorithm}
% \caption{Adversarial Training Algorithm for Solving Nash Equilibrium in Zero-Sum Games }
% \begin{algorithmic}[1] 
% \State \textbf{Initialize: } Parameters $\theta_f,\theta_g$ for policy models and $\theta_v$ for state value model.
% \State Set $\theta^{\text{target}}_f \leftarrow \theta_f$, $\theta^{\text{target}}_g \leftarrow \theta_g$, and $\theta^{\text{target}}_v \leftarrow \theta_v$ .

% \For{$i = 1, 2, 3, .... $}

% \State Randomly initialize the starting point $s_0$.
% \State Run $f_{\theta^{\text{target}}_f}(a|s),\text{ } g_{\theta^{\text{target}}_g}(a|s)$ for $T$ steps
% \State Calculate estimated advantages for player $g$: $\hat{A}^f_0,\hat{A}^f_1,...,\hat{A}^f_{T-1}$

% \State Optimize surrogate loss function $L^{g,\text{CLIP}}$ w.r.t. $\theta_{g}$ \Comment{$g$ step}
% \State Optimize state value estimation loss $L^v$ w.r.t. $\theta_v$

% \If{$i\text{ } \% \text{ }K_{\text{f}} = 0  $} \Comment{$f$ step}
% \State Optimize surrogate loss function $L^{f,\text{CLIP}}$ w.r.t. $\theta_f$
% \EndIf

% \If{$i\text{ } \% K_{\text{refresh}}\text{ } = 0  $} \Comment{refresh target networks}
% \State Set $ \theta_f \leftarrow \theta^{\text{target}}_f$, $\theta_{g}  \leftarrow \theta^{\text{target}}_{g}$, and $\theta_v \leftarrow  \theta^{\text{target}}_v $ .
% \EndIf

% \EndFor

% \end{algorithmic}
% \end{algorithm}

\subsection{Analysis}

For solving Nash Equilibrium in zero-sum discounted stochastic games, we train an agent to always compete against its best possible opponent. We can think of the training process under objective function (3) as performing gradient ascent for 
$$F(f) = \min_{g} \frac{1}{N_s}\sum_{s \in \mathcal{S}} v^{f,g}(s).$$

In this section, we consider the case as if we use a tabular approach for each agent's policy instead of using model approximations. The tabular approach means that for any $s$ and any $a^f,a^g$, $f(a^f|s)$ or $g(a^g|s)$ would be variables, and the only constraints are standard probability requirements. We show that using \eqref{obj_f} (or \eqref{obj_g}) as the objective function to solve for $f^*$ (or $g^*$) is theoretically a sound choice, because there is no local maximum in $F(f)$. 

\textbf{Proposition 1: } Function $F(f)$ has no local maxima. Namely, if at a certain $\tilde{f}$ there exists no feasible strictly ascent direction for $F(\tilde{f})$, then $$F(\tilde{f}) = \max_f F(f)\text{ } \text{ s.t. } \bm{f}(s)\geq 0,\bm{f}_s^T\bm{1} = 1 \text{ } \forall s \in \mathcal{S}. $$

\begin{proof}

%\textbf{Proof: }

Without loss of generality we can assume that there exists a state $s_0$ under which both agents receive zero rewards, and regardless of the actions the agents take, any $s \in \mathcal{S}$ has the same probability to be visited at the next step. Formally, 
\begin{align}
R(s_0) & = 0, \nonumber \\
p(s^\prime|s_0,a^f,a^g) &= \frac{1}{N_s} \text{ for }\forall s^\prime \in \mathcal{S},\forall a^f \in \mathcal{A}_f, \forall a^g \in \mathcal{A}_g. \label{dummy_state_positive_prob}
\end{align}
Given policies $f,g$ and the discount factor $\gamma \in (0,1)$, we have
$ v^{f,g}(s_0) = \frac{1}{\gamma N_s}\sum_{s\in\mathcal{S}}v^{f,g}(s).$ Therefore,
$$F(f) = \frac{1}{\gamma} \min_g v^{f,g}(s_0). $$
We hereby use this artificial starting state $s_0$ to simplify the notation. The conclusions below are adapted to the multi-agent setting in our case from the reinforcement learning perspective.

The policy gradient theorem (a variant of (13.5) in \cite{sutton1998reinforcement})  where the agents employ policies $f$ and $g$, the starting state is $s_0$, and policy $f$ is a function $f_{\theta}$ parametrized by $\theta$ reads
%\begin{equation}
\begin{align}
\nabla_{\theta} v^{f,g}_f(s_0) &= \sum_s d^{f,g}(s) \sum_{a^f} Q^{f,g}_f(s,a^f)\nabla_{\theta} f(a^f|s)   \nonumber \\
 &= \sum_s d^{f,g}(s) \sum_{a^f} A^{f,g}_f(s,a^f)\nabla_{\theta} f(a^f|s) + \sum_s d^{f,g}(s) \sum_{a^f} v^{f,g}(s)\nabla_{\theta} f(a^f|s)  \nonumber\\
 &= \sum_s d^{f,g}(s) \sum_{a^f} A^{f,g}_f(s,a^f)\nabla_{\theta} f(a^f|s). \nonumber
\end{align}
%\end{equation}
Note that the last equality holds because $\sum_{a^f} f(a|s) = 1$ implies that $\sum_{a^f}\nabla_{\theta} f(a^f|s) = 0$. Here $d^{f,g}(s)$ is the expected frequency of encountering state $s$ discounted by $\gamma$. In other words, the performance of the agent is differentiable, and the current advantage function can be used to characterize the gradient. Particularly, under the tabular representation, a policy at each state $s$ is
$$\bm{f}_s = (f_{s,1},f_{s,2},\ldots,f_{s,|\mathcal{A}_f|-1}, 1 - \sum_{i=1}^{|\mathcal{A}_f|-1}f_{s,i}) $$
where the parameter $\theta$ for policy model $f$ are variables $f_{s,1},f_{s,2},\ldots,f_{s,|\mathcal{A}_f|-1}$ at $s \in \mathcal{S}$. Therefore, for $i = 1,2,\ldots,|\mathcal{A}_f|$,
$$\frac{\partial v^{f,g}(s_0)}{\partial f_{s,i}} = d^{f,g}(s)\Big( A^{f,g}_f\big(s,a^f(i)\big) -  A^{f,g}_f\big(s,a^f(|\mathcal{A}_f|)\big)  \Big).$$
Due to \eqref{dummy_state_positive_prob}, we know that there is a positive probability to visit any $ s \in \mathcal{S}$ when starting from $s_0$. Therefore,
\begin{align}
d^{f,g}(s) > 0 \text{ }\text{ for } s \in \mathcal{S}. \label{d_positive}
\end{align}

Based on optimality condition (3.17) in \cite{sutton1998reinforcement} when competing against a policy $f$, for optimal policy $g^*$ satisfies
\begin{align}
 v^{f,g^*}(s) & =  -\max_{a^g} Q^{f,g^*}_g(s,a^g), \nonumber \\
Q^{f,g^*}_g(s,a^g) & = \sum_{s^\prime \in \mathcal{S}, a^f \sim f(a|s) }p(s^\prime|s,a^f,a^g)\Big[-R(s) + \gamma \max_{a^{\prime g}}Q^{f,g^*}_g(s^\prime,a^{\prime g})\Big], \nonumber 
\end{align}
for any $s \in \mathcal{S}$. Similarly, when playing against a policy $g$, optimal policy $f^*$ satisfies
\begin{align}
 v^{f^*,g}(s) & =  \max_{a^f} Q^{f^*,g}_f(s,a^f), \nonumber \\
Q^{f^*,g}_f(s,a^f) & = \sum_{s^\prime \in \mathcal{S}, a^g \sim g(a|s) }p(s^\prime|s,a^f,a^g)\Big[R(s) + \gamma \max_{a^{\prime f}}Q^{f^*,g}_f(s^\prime,a^{\prime f})\Big], \nonumber 
\end{align}

% In a MDP with a static environment, the optimal policy $\pi^*$ for the agent satisfies that, for any $s$,
% %\begin{equation}
% \begin{align}
%  v^{\pi^*}(s) & =  \max_a Q^{\pi^*}(s,a) \nonumber \\
% \text{where } Q^{\pi^*}(s,a) & = \sum_{s^\prime}p(s^\prime|s,a)\Big[R(s) + \gamma \max_{a^\prime}Q^{\pi^*}(s^\prime,a^\prime)\Big]. \nonumber 
% \end{align}
% %\end{equation}
for any $s \in \mathcal{S}$. 

The policy improvement theorem (inequalities (4.7) and (4.8) in \cite{sutton1998reinforcement}) when playing against a certain policy $g$, for two policies $f$ and $f^\prime$, states that if
$$v^{f,g}(s) \leq Q^{f,g}_f(s,f^\prime(a|s)) \triangleq \mathbb{E}_{a^f \sim f^\prime(a|s)} Q^{f,g}_f(s,a^f)   $$
holds for any state $s$, then
$$v^{f^\prime,g}(s) \geq v^{f,g}(s)\text{ }\text{ for } s \in \mathcal{S}. $$
Similarly, if 
$$v^{f,g}(s) \geq Q^{f,g}_g(s,g^\prime(a|s)) \triangleq \mathbb{E}_{a^g \sim g^\prime(a|s)} Q^{f,g}_g(s,a^g)   $$
holds for any state $s$, then
$$v^{f,g^\prime}(s) \leq v^{f,g}(s)\text{ }\text{ for } s \in \mathcal{S}. $$

Although the policy improvement theorem was initially established for pure policies, the line of logic in its proof also holds for mixed policies.

% if for two policies $\pi$ and $\pi^\prime$, 
% $$v^{\pi}(s) \leq Q^{f,g}_f(s,f^\prime(a|s)) \triangleq \mathbb{E}_{a^f \sim f^\prime(a|s)} Q^{f,g}_f(s,a^f)   $$
% holds for any state $s$, then
% $$v^{\pi^\prime}(s) \geq v^{\pi}(s)\text{ }\text{ }\text{ }\text{ for }\forall s \in \mathcal{S}. $$
% Note that, although the theorem was initially established for pure strategies, the line of logic in its proof also holds for mixed strategies and the conclusion can be generalized for mixed strategies.

Let us define set $ g_P $ to be the set containing any possible pure policy for agent $g$. Then it is clear that
$$ \min_g v^{f,g}(s_0) = \min_{g \in g_P} v^{f,g}(s_0).$$
% Otherwise, we assume the existence a mixed policy $\tilde{g}$ such that
% \begin{align}
% v^{f,\tilde{g}}(s_0) < \min_{g \in g_P} v^{f,g}(s_0).\label{mixed_g_smaller}
% \end{align}
% From $\tilde{g}$, we construct a pure policy $\tilde{g}_P$ with
% \[
%     \tilde{g}_P\big(a^g(i)|s\big)= 
% \begin{cases}
%     1 & \text{if } i = \min_k G_s\\
%     0              & \text{otherwise}.
% \end{cases}
% \]
% where $G_s = \Big\{k\text{ }\Big|\text{ }Q^{f,\tilde{g}}_g\big(s,a^g(k)\big) = \min_j Q^{f,\tilde{g}}_g\big(s,a^g(j)\big)\Big\}$ is the index set of optimal actions of policy $\tilde{g}$ at state $s$. Pure policy $\tilde{g}_P$ plays only the optimal action, and if multiple optimal actions exist at a certain state, it chooses the one with the smallest index. We thus have
% $$Q^{f,\tilde{g}}_g\big(s,\tilde{g}_P(a|s)\big)\leq v^{f,\tilde{g}}(s)\text{ for }\forall s, $$
% and with policy improvement theorem we have
% \begin{align}
% v^{f,\tilde{g}_P}(s) \leq v^{f,\tilde{g}}(s) \text{ for } \forall s. \label{mixed_g_not_smallest}
% \end{align}
% Since \eqref{mixed_g_smaller} contradicts \eqref{mixed_g_not_smallest}, we have
% $$F(f) = \frac{1}{\gamma} \min_g v^{f,g}(s_0) = \frac{1}{\gamma} \min_{g \in g_P} v^{f,g}(s_0).$$

We now switch to the main part of the proof, which includes a few claims listed below.

For a given $f$, we consider the performance of $g$ on the set $g_P$. We evaluate $v^{f,g}(s_0)$ for every $g\in g_P$, which yields the set $\big\{v^{f,g_{P_1}}(s_0),v^{f,g_{P_2}}(s_2),\ldots,v^{f,g_{P_K}}(s_0)\big\}$ of distinct values. We rank those values in the ascending order to get an ordered set $$\big\{v^{(1)}(f),\text{ }v^{(2)}(f),\text{ }\ldots\big\} .$$
Here $v^{(1)}(f) = \min_{g \in g_P} v^{f,g}(s_0)$, $v^{(2)}(f)$ is the second smallest value, and so on. We have
\begin{align}
\delta(f) = v^{(2)}(f) - v^{(1)}(f) > 0 \label{delta_f_positive}
\end{align}
which is a strictly positive gap between the first and second smallest performance on set $g_P$. We also define the best response set
$$g^{\text{best}}_P(f) = \{g \in g_P\text{ }|\text{ }v^{f,g}(s_0) = v^{(1)}(f) \}. $$

\textbf{Claim 1: } For a policy vector $\bm{f} = (\bm{f}_{s_1}^T,\bm{f}_{s_2}^T,\ldots,\bm{f}_{s_{N_s}}^T)^T$ where $\bm{f}_{s_i} \in \mathbb{R}^{|\mathcal{A}_f|}$ is the policy vector at state $s_i$, a vector $\Delta_f = (\Delta_{f,1},\Delta_{f,2},\ldots,\Delta_{f,s_i})^T$ where $\Delta_{f,i} \in \mathbb{R}^{|\mathcal{A}_f|}$ and $\Delta_{f,i}^T\bm{1} = 0$ for $i=1,2,\ldots,N_s$, and an $\epsilon_1>0$ such that
\begin{align}
\bm{f} + \epsilon_1 \Delta_f \geq 0, \label{claim1}
\end{align}
we have 
$$g^{\text{best}}_P(\bm{f}+\epsilon\Delta_f)\subseteq g^{\text{best}}_P(f)$$
for any $\epsilon>0$ small enough.

\begin{proof}

Because of the policy gradient theorem and the fact that $R$ is bounded due to the finite state space $\mathcal{S}$, there exists $M>0$ such that 
$$\Big|\frac{\partial v^{\tilde{f},\tilde{g}}(s_0) }{\partial \tilde{f}_{s,i}}\Big|<M \text{  }\forall \tilde{f},\tilde{g},s \in \mathcal{S},i=1,2,\ldots,|\mathcal{A}_f|-1.  $$

By the assumption \eqref{claim1} of the claim, there exists a small enough $\epsilon_2 > 0$ such that for $0 < \epsilon < \min(\epsilon_1,\epsilon_2,\frac{\delta(f)}{2M})$, we have $\bm{0} \leq \bm{f}+\epsilon\Delta_f \leq \bm{1}$ (so it is still a well-defined policy) and, by the definition of the derivative and the definition of $M$, for any $g\in g_P$ we have
$$\big|v^{\bm{f}+\epsilon\Delta_f,g}(s_0) - v^{\bm{f},g}(s_0) \big| < \frac{\delta(f)}{2}.$$

Thus, for every $g \in g^{\text{best}}_P(f)$, since $v^{(1)}(f) = v^{f,g}(s_0)$, we have
$$v^{\bm{f}+\epsilon\Delta_f,g}(s_0) < v^{(1)}(f) + \frac{\delta(f)}{2}.$$
And for every $\tilde{g} \in g_P \setminus g^{\text{best}}_P(f)$, we have
$$v^{\bm{f}+\epsilon\Delta_f,\tilde{g}}(s_0) > v^{(2)}(f) - \frac{\delta(f)}{2} = v^{(1)}(f) + \frac{\delta(f)}{2}.$$
Therefore, for every $g \in g^{\text{best}}_P(f)$, we have  $\tilde{g}\not\in \operatorname*{argmin}_{g} v^{\bm{f}+\epsilon\Delta_f,g}(s_0)$, which implies that given the feasible direction $\Delta_f$ for policy $f$ with a corresponding $\epsilon_1>0$, for any small enough $\epsilon>0$ we have
\begin{align}
g^{\text{best}}_P(\bm{f}+\epsilon\Delta_f)\subseteq g^{\text{best}}_P(f). \label{best_g_neighbor}
\end{align}
This shows the claim.
\end{proof}

% Note that in the extreme case where every $g \in g_P$ yields exactly the same performance $v^{f,g}(s_0)$, though $v^{(2)}(f)$ no longer exists, it is obvious that $g^{\text{best}}_P(f) = g_P$ so \eqref{best_g_neighbor} still holds.

\textbf{Claim 2: } Let $f$ be a local maximum for $F$. Then for any $s \in \mathcal{S}$, the linear system
\begin{equation} \label{LS_original}
\begin{gathered}
\Delta^T  \bm{A}_s > \bm{0} \\
\bm{f}_s + \Delta \geq \bm{0} \\
(\bm{f}_s + \Delta)^T \bm{1} = 1
\end{gathered}
\end{equation}
is infeasible with $\Delta \in \mathbb{R}^{|\mathcal{A}_f|}$ as variables, where the advantage matrix is
$$\bm{A}_s = \Big( A^{f,g_{P_j}}\big(s,a^f(i)\big) \Big)_{i,j}$$
with $g_{P_j} \in g^{\text{best}}_P(f)\text{ for }j=1,2,\ldots,|g^{\text{best}}_P(f)|.$ 

\begin{proof}

We show the statement by contradiction. Let us assume the existence of a state $s \in \mathcal{S}$ for which the linear system \eqref{LS_original} is feasible with $\Delta = (\Delta_1,\Delta_2,\ldots,\Delta_{|\mathcal{A}_f|})^T \in \mathbb{R}^{|\mathcal{A}_f|}$. This implies that the conditions of Claim 1 are met and thus $g^{\text{best}}_P(\bm{f}+\epsilon\Delta_f)\subseteq g^{\text{best}}_P(f)$ for any small enough $\epsilon>0$. Here $\Delta_f = (\bm{0}^T,\bm{0}^T,\ldots,\Delta^T,\ldots,\bm{0}^T)^T$ with $\Delta$ being at the position corresponding to state $s$. Due to $\bm{f}_s^T\bm{1} = 1$, it is also obvious that $\Delta^T\bm{1} = 0$ and we have
$$\Delta_{|\mathcal{A}_f|} = - \sum_{i=1}^{|\mathcal{A}_f|-1}\Delta_i.$$

Observe that each row of $\bm{A}_s$ corresponds to an action in $\mathcal{A}_f$, and each column of $\bm{A}_s$ corresponds to a optimal pure policy for $g$ when playing against the given $f$. This suggests that for $j = 1,2,\ldots,|g^{\text{best}}_P(f)|$ we have,
$$\sum_{i=1}^{|\mathcal{A}_f|-1}\Delta_i\Big( A^{f,g_{P_j}}\big(s,a^f(i)\big) - A^{f,g_{P_j}}\big(s,a^f(|\mathcal{A}_f|)\big)  \Big) > 0. $$
By the policy gradient theorem, for $j = 1,2,\ldots,|g^{\text{best}}_P(f)|$ we have
\begin{align}
&\text{ } \sum_{i=1}^{|\mathcal{A}_f|-1} \frac{\partial v^{f,g_{P_j}}(s_0)}{f_{s,i}}\Delta_i \nonumber \\
& = d^{f,g_{P_j}}(s)\sum_{i=1}^{|\mathcal{A}_f|-1}\Delta_i\Big( A^{f,g_{P_j}}\big(s,a^f(i)\big) - A^{f,g_{P_j}}\big(s,a^f(|\mathcal{A}_f|)\big)  \Big) \nonumber \\
& >0. \label{gradient_lg_0}
\end{align}
The last inequality holds strictly since we have already argued that $d^{f,g}(s)>0 \text{ for every }s,f,g$ when the starting point is the artificial state $s_0$. Inequality \eqref{gradient_lg_0} shows that for any $g \in g^{\text{best}}_P(f)$, the directional gradient for $v$ is strictly positive along $\Delta$. 

Therefore, there exists $\epsilon_3 >0$ such that for $0 < \epsilon < \epsilon_3$ we have,
\begin{align}
v^{\bm{f}_s + \epsilon\Delta, g_{P_j}}(s_0) > v^{f, g_{P_j}}(s_0) \text{ for }j = 1,2,\ldots,|g^{\text{best}}_P(f)|. \label{f_improve_neighbor}
\end{align}

Given Claim 1 and \eqref{f_improve_neighbor}, we know that if at a certain state $s$ there exists a vector $\Delta$ feasible to \eqref{LS_original}, then there exists $\tilde{\epsilon}>0$ such that for any $0 < \epsilon < \tilde{\epsilon}$ we have,
\begin{align}
F(\bm{f}_s + \epsilon\Delta) & = \frac{1}{\gamma}\min_{g\in g_P}v^{\bm{f}_s + \epsilon\Delta,g}(s_0) \nonumber \\
& = \frac{1}{\gamma}\min_{g\in g^{\text{best}}_P(\bm{f}_s + \epsilon\Delta)}v^{\bm{f}_s + \epsilon\Delta,g}(s_0) \nonumber \\
& \geq \frac{1}{\gamma}  \min_{g\in g^{\text{best}}_P(f)}v^{\bm{f}_s + \epsilon\Delta,g}(s_0)\text{ }\text{ }\text{  (due to \eqref{best_g_neighbor})} \nonumber \\
&> \frac{1}{\gamma}  \min_{g\in g^{\text{best}}_P(f)}v^{f,g}(s_0)\text{ }\text{ }\text{  (due to \eqref{f_improve_neighbor})} \nonumber \\
 &=F(f). \nonumber
\end{align}

We conclude that $f$ can not be a local maximum of $F(f)$ as long as there exists a state $s$ such that \eqref{LS_original} is feasible. 
\end{proof}

\textbf{Claim 3: } If $f$ is a local maximum for $F$, then for every state $s$ there exists a vector $\bm{w}_s$ such that 
$$\Delta^T\bm{A}_s\bm{w}_s \leq 0,\bm{w}_s \geq 0, \bm{w}_s^T\bm{1} = 1,  $$
for any vector $\Delta$ that makes $\bm{f}_s + \Delta$ a well-defined policy at state $s$.

\begin{proof}

First, we reorder the actions of $f$ so that the one with the highest probability at state $s$ is the last one in $\bm{f}_s$. Then, consider
\begin{equation} \label{LS_simplified}
\begin{gathered}
\tilde{\Delta}^T \tilde{\bm{A}}_s > 0 \\
\tilde{\Delta}_i \geq 0 \text{ }\text{ } \forall i \in \mathcal{C}_s \\
\end{gathered}
\end{equation}
where index set $\mathcal{C}_s = \{i\leq|\mathcal{A}_f|-1\text{ }|\text{ }f_{s,i} = 0\}$, vector $\tilde{\Delta} \in \mathbb{R}^{|\mathcal{A}_f|-1}$, and
$$\tilde{\bm{A}}_s = \Big(\bm{A}^T_{s,1} - \bm{A}^T_{s,|\mathcal{A}_f|},\bm{A}^T_{s,2} - \bm{A}^T_{s,|\mathcal{A}_f|},\ldots,\bm{A}^T_{s,|\mathcal{A}_f|-1} - \bm{A}^T_{s,|\mathcal{A}_f|}\Big)^T $$
with $\bm{A}_{s,i}$ being the $i$-th row vector of $\bm{A}_s$.

By Claim 2, we know that \eqref{LS_original} is infeasible. We now argue that if \eqref{LS_original} is infeasible, so is \eqref{LS_simplified}. Let us assume that \eqref{LS_simplified} has a solution $\tilde{\Delta}$. Then there exists a small enough $\epsilon > 0$ such that $f_{s,i} + \epsilon\tilde{\Delta}_i \geq 0$ for all $i = 1,2,\ldots,|\mathcal{A}_f|-1$. This is true since $f_{s,i} = 0, \tilde{\Delta}_i \geq 0$ holds for every $i \in \mathcal{C}_s$, and for $i \not\in \mathcal{C}_s$ we have $f_{s,i}>0$ and thus the inequality holds for small enough $\epsilon > 0$. We denote one such appropriate value as $\epsilon_1$.
Finally, if we let $\Delta = (\tilde{\Delta}^T, -\tilde{\Delta}^T\bm{1})^T$, then we clearly have $\Delta^T\bm{1} = 0$, and $\bm{f}_s + \epsilon\Delta \geq \bm{0}$ for any $\epsilon$ with
$$0 < \epsilon < \epsilon_2 = \min\Big( \epsilon_1, \max( 0, \frac{ f_{s,|\mathcal{A}_f|} }{\tilde{\Delta}^T\bm{1} } ) \Big) .$$
Since $f_{s,|\mathcal{A}_f|}>0$, we have $\epsilon_2 > 0.$ It is also easy to check that
$$\Delta^T\bm{A}_s  = \epsilon\tilde{\Delta}^T \tilde{\bm{A}}_s > 0.$$
We therefore conclude that if $f$ is a local maximum, then \eqref{LS_simplified} is infeasible. 

In \eqref{LS_simplified}, except for the strict inequality let us denote all other constraints as $\bm{B}\tilde{\Delta}\geq 0$. Note that $\bm{B}$ is disposed with 0,1 on the diagonal. By the theorem of alternatives, infeasibility of \eqref{LS_simplified} implies that there exist $\bm{y}\geq 0, \bm{z}\geq 0 $ so that
$$\tilde{\bm{A}}_s \bm{y} + \bm{B}^T \bm{z} = 0 \text{ and } \bm{y} \neq 0.$$

After rescaling $\bm{y}$, there exist $k>0$ and column vector $\bm{w}_s\geq 0$ with $\bm{w}^T_s\bm{1} = 1$ such that
$$\tilde{\bm{A}}_s \bm{w}_s = -k\bm{B}^T \bm{z}.$$
Then, due to $\bm{B}\tilde{\Delta} \geq \bm{0}$ and $\bm{B}_{u,v}=0 \text{ for } u \not\in \mathcal{C}_s, v \not\in \mathcal{C}_s$, for any $\tilde{\Delta}$ with $\tilde{\Delta}_{i} \geq 0$ for $i \in \mathcal{C}_s$, we have
\begin{align}
\tilde{\Delta}^T\tilde{\bm{A}}_s w_s = -k\tilde{\Delta}^T\bm{B}^T \bm{z}\leq 0. \label{tilde_delta_feasible_leq_0}
\end{align}

We define a new vector $\Delta$ such that $\bm{f}_s + \Delta \geq \bm{0}, \Delta^T\bm{1} = 0$. Note that these are equivalent to saying that $\bm{f}_s + \Delta$ is a policy. We have $\Delta_i \geq 0 \text{ for } i \in \mathcal{C}_s$ and $ \Delta_{|\mathcal{A}_f|} = - \sum_{i=1}^{|\mathcal{A}_f|-1}\Delta_i$. If we let
$$\tilde{\Delta} = (\Delta_1, \Delta_2,\ldots,\Delta_{|\mathcal{A}_f|-1})^T,$$
then we have 
\begin{align}
\tilde{\Delta}^T\tilde{\bm{A}}_s \bm{w}_s &= ( \Delta_1, \Delta_2,\ldots,\Delta_{|\mathcal{A}_f|-1} ) \Big(\bm{A}^T_{s,1} - \bm{A}^T_{s,|\mathcal{A}_f|},\bm{A}^T_{s,2} - \bm{A}^T_{s,|\mathcal{A}_f|},\ldots,\bm{A}^T_{s,|\mathcal{A}_f|-1} - \bm{A}^T_{s,|\mathcal{A}_f|}\Big)^T \bm{w}_s \nonumber \\
& =\Big( \sum_{i=1}^{|\mathcal{A}_f|-1}\Delta_i\bm{A}_{s,i} - \sum_{i=1}^{|\mathcal{A}_f|-1}\Delta_i\bm{A}_{s,|\mathcal{A}_f|} \Big)\bm{w}_s \nonumber \\
& = (\Delta_1, \Delta_2,\ldots,\Delta_{|\mathcal{A}_f|-1},-\sum_{i=1}^{|\mathcal{A}_f|-1}\Delta_i)\Big(\bm{A}_{s,1}^T,\bm{A}_{s,2}^T,\ldots,\bm{A}_{s,|\mathcal{A}_f|}^T\Big)^T\bm{w}_s   \nonumber \\
& = \Delta^T \bm{A}_s \bm{w}_s. \nonumber
\end{align}
Together with \eqref{tilde_delta_feasible_leq_0}, we thus have
\begin{align}
\Delta^T \bm{A}_s \bm{w}_s \leq 0, \label{LS_adv_leq_0}
\end{align}
which shows the claim.
\end{proof}

% $$\tilde{\Delta}^T\tilde{\bm{A}}_s \bm{w}_s = ( \Delta_1, \Delta_2,\ldots,\Delta_{|\mathcal{A}_f|-1} ) \Big(\bm{A}^T_{s,1} - \bm{A}^T_{s,|\mathcal{A}_f|},\bm{A}^T_{s,2} - \bm{A}^T_{s,|\mathcal{A}_f|},\ldots,\bm{A}^T_{s,|\mathcal{A}_f|-1} - \bm{A}^T_{s,|\mathcal{A}_f|}\Big)^T  $$
% according to \eqref{tilde_delta_feasible_leq_0},
% which implies that for any feasible $\Delta$ such that $\bm{f}_s + \Delta \geq \bm{0}, \Delta^T\bm{1} = 0$, we have
% \begin{align}
% \Delta^T \bm{A}_s \bm{w}_s \leq 0. \label{LS_adv_leq_0}
% \end{align}

Let $f$ be a local maximum of $F$. Treating $\bm{w}_s$ defined in Claim 3 as coefficients for a convex combination of opponent's policies in set $g^{\text{best}}_P(f) = \{g^{f,1}_{P},g^{f,2}_{P},\ldots,g^{f,|g^{\text{best}}_P(f)|}_{P}\}$ (namely, the probability that the agent plays the $g^{f,j}_{P}$ at state $s$ is equal to the $j$-th element of $\bm{w}_s$), we obtain a mixed policy $\tilde{g}^w_s$ at state $s$ for the opponent agent $g$, the advantage function of which is $A^{f,\tilde{g}^w_s}_f\Big(s,a^f(i)\Big) = \big(\bm{A}_s\bm{w}_s\big)_i$ for action $a^f(i)$. Regarding this advantage function, we establish that
\begin{align}
 v^{f,\tilde{g}^w_s}(s) &= \sum_{i=1}^{|\mathcal{A}_f|} f_{s,i}\text{ } Q^{f,\tilde{g}^w_s}_f\Big(s,a^f(i)\Big) \text{ for }\ s \in \mathcal{S},\text{ implies that} \nonumber \\
0 &= \sum_{i=1}^{|\mathcal{A}_f|} f_{s,i}\bigg( Q^{f,\tilde{g}^w_s}_f\Big(s,a^f(i)\Big) - v^{f,\tilde{g}^w_s}(s)  \bigg) \text{ for } s \in \mathcal{S},\text{ and thus} \nonumber \\
0 & = \sum_{i=1}^{|\mathcal{A}_f|} f_{s,i}\text{ }A^{f,\tilde{g}^w_s}_f\Big(s,a^f(i)\Big)\text{ for } s \in \mathcal{S}, \text{ and}\nonumber \\
0 & = \sum_{1 \leq i \leq |\mathcal{A}_f|, i \not\in \mathcal{C}_s} f_{s,i}\text{ }A^{f,\tilde{g}^w_s}_f\Big(s,a^f(i)\Big)\text{ for }s \in \mathcal{S}. \label{LS_sum_adv_leq_0}
\end{align}
Since $\bm{f}_s$ is a well-defined distribution, the set $\mathcal{C}_s^c = \{i\in\mathbb{N}\text{ }|\text{ }1 \leq i \leq |\mathcal{A}_f|, f_{s,i}>0\}$ is non-empty. With \eqref{LS_sum_adv_leq_0} and $\bm{f}_s \geq \bm{0}$, we see that there exist $j_1,j_2 \in \mathcal{C}_s^c$ ($j_1$ and $j_2$ can be identical) such that
\begin{align}
A^{f,\tilde{g}^w_s}_f\Big(s,a^f(j_1)\Big) & \leq 0,\text{ } f_{s,j_1} > 0, \label{j1} \\
A^{f,\tilde{g}^w_s}_f\Big(s,a^f(j_2)\Big) & \geq 0,\text{ } f_{s,j_2} > 0. \label{j2}
\end{align}

\textbf{Claim 4: }We have
\begin{align}
A^{f,\tilde{g}^w_s}_f\Big(s,a^f(i)\Big) \leq 0 \text{ for } \forall\text{ } a^f(i). \label{char_adv_1}
\end{align}

\begin{proof}

If \eqref{char_adv_1} does not hold, then there exists an index $i_1$ such that $A^{f,\tilde{g}^w_s}_f\Big(s,a^f(i_1)\Big) = \big(\bm{A}_s\bm{w}_s\big)_{i_1} > 0$. Let $\Delta^1$ be the vector defined as
\[
    \Delta^1_l= 
\begin{cases}
    1 & \text{if } l=i_1\\
    -1 & \text{if } l=j_1\\
    0              & \text{otherwise}.
\end{cases}
\]
(Note that $f_{s,i_1}<1$, since otherwise, $\mathcal{C}_s^c$ contains only $i_1$, and \eqref{LS_sum_adv_leq_0} implies $A^{f,\tilde{g}^w_s}_f\Big(s,a^f(i_1)\Big) = 0$. )

Meanwhile, given \eqref{j1} we have $f_{s,j_1}>0$. Therefore, there exists a $0< \epsilon_{(1)} < \min(1 - f_{s,i_1},\text{ } f_{s,j_1})$  such that
\begin{align}
\bm{0} & \leq \bm{f}_s + \epsilon_{(1)} \Delta^1 \nonumber \\
& = \Big( f_{s,1},f_{s,2},\ldots,f_{s,i_1} + \epsilon_{(1)},\ldots,f_{s,j_1} - \epsilon_{(1)},\ldots,f_{s,|\mathcal{A}_f|} \Big) \nonumber \\
& \leq \bm{1}, \nonumber
\end{align}
and 
$$\epsilon_{(1)}(\Delta^{1})^T\bm{A}_s\bm{w}_s = \epsilon_{(1)} \bigg( A^{f,\tilde{g}^w_s}_f\Big(s,a^f(i_1)\Big) - A^{f,\tilde{g}^w_s}_f\Big(s,a^f(j_1)\Big)  \bigg) > 0,$$
which contradicts Claim 3. Therefore, such $i_1$ does not exist, and advantage for any action $a^f(i)$ is non-positive.
\end{proof}

\textbf{Claim 5: }We have
\begin{align}
A^{f,\tilde{g}^w_s}_f\Big(s,a^f(i)\Big) = 0 \text{ for } \forall\text{ } a^f(i) \text{ with }f_{s,i}>0. \label{char_adv_2}
\end{align}

\begin{proof}

If \eqref{char_adv_2} does not hold, then there exists an index $i_2$ such that $A^{f,\tilde{g}^w_s}_f\Big(s,a^f(i_2)\Big) = \big(\bm{A}_s\bm{w}_s\big)_{i_2} < 0$ and $f_{s,i_2}>0$. Let $\Delta^2$ be the vector defined as
\[
    \Delta^2_l= 
\begin{cases}
    -1 & \text{if } l=i_2\\
    1 & \text{if } l=j_2\\
    0              & \text{otherwise}.
\end{cases}
\]
(Note that $f_{s,j_2}<1$ because we already have $f_{s,i_2}>0$.)
Therefore, there exists a $0< \epsilon_{(2)} < \min(1 - f_{s,j_2},\text{ } f_{s,i_2})$  such that
\begin{align}
\bm{0} & \leq \bm{f}_s + \epsilon_{(2)} \Delta^2 \nonumber \\
& = \Big( f_{s,1},f_{s,2},\ldots,f_{s,i_2} - \epsilon_{(2)},\ldots,f_{s,j_2} + \epsilon_{(2)},\ldots,f_{s,|\mathcal{A}_f|} \Big) \nonumber \\
& \leq \bm{1}, \nonumber
\end{align}
and 
$$\epsilon_{(2)}(\Delta^{2})^T\bm{A}_s\bm{w}_s = \epsilon_{(2)} \bigg( A^{f,\tilde{g}^w_s}_f\Big(s,a^f(j_2)\Big) - A^{f,\tilde{g}^w_s}_f\Big(s,a^f(i_2)\Big)  \bigg) > 0,$$
which contradicts Claim 3. Therefore, such $i_2$ does not exist, and advantage is zero for any action $a^f(i)$ with $f_{s,i}>0$.
\end{proof}

In summary, the claims show that when policy $f$ is a local maximum, we obtain a mixed policy $\tilde{g}^w_s$ for any state $s$. We denote the entire policy function as $\tilde{g}^w$. Obviously, $\tilde{g}^w$ only plays optimal actions against $f$ and is also a best response to $f$. Meanwhile, under this policy $\tilde{g}^w$, due to \eqref{char_adv_1} and \eqref{char_adv_2}, at any state $s$ we have
\begin{equation} 
\begin{gathered}
A^{f,\tilde{g}^w_s}_f\Big(s,a^f(i)\Big) \leq 0 \text{ for } i=1,2,\ldots,|\mathcal{A}_f|, \nonumber \\
A^{f,\tilde{g}^w_s}_f\Big(s,a^f(i)\Big) = 0 \text{ if } f_{s,i}>0.  \\
\end{gathered}
\end{equation}
Therefore, $f$ is also the best response to $\tilde{g}^w$ as prescribed by the Bellman optimality condition
$$v^{f,\tilde{g}^w}(s) = \max_{a^f \in \mathcal{A}_f} Q^{f,\tilde{g}^w}_f(s,a^f) \text{ }\text{ }s \in \mathcal{S}.$$

We use this to show that $f$ is the global maximum of function $F$. Since $f$ is the best response to $\tilde{g}^w$, for any policy $\tilde{f}$, under $\tilde{g}^w$ we see that
\begin{align}
v^{f,\tilde{g}^w}(s) & = \max_{a^f \in \mathcal{A}_f} Q^{f,\tilde{g}^w}_f(s,a^f) \nonumber \\
& \geq \mathbb{E}_{a^f \sim \tilde{f}(a|s)}Q^{f,\tilde{g}^w}_f(s,a^f) \nonumber \\
& = Q^{f,\tilde{g}^w}_f(s,\tilde{f}(a|s)) \nonumber
\end{align}
holds for any state $s$. By the policy improvement theorem, we thus have
$$v^{f,\tilde{g}^w}(s) \geq v^{\tilde{f},\tilde{g}^w}(s)\text{ } \text{ } \text{for } \forall s.$$

Since $\tilde{g}^w$ is a best response to $f$, we know that
$$F(f) = \frac{1}{N_s}\sum_{s \in \mathcal{S}} v^{\tilde{f},\tilde{g}^w}(s).$$
Therefore,
\begin{equation} 
\begin{gathered}
F(\tilde{f}) \leq \frac{1}{N_s}\sum_{s \in \mathcal{S}}v^{\tilde{f},\tilde{g}^w}(s) \leq \frac{1}{N_s}\sum_{s \in \mathcal{S}}v^{f,\tilde{g}^w}(s)= F(f). \nonumber
\end{gathered}
\end{equation}
This concludes the proof.
\end{proof}

\section{Experimental Study}

In this section, we illustrate the proposed IRL and Nash Equilibrium algorithms on a zero-sum stochastic game and present their performances. First, we introduce the zero-sum stochastic game we use in our experiments, and discuss the complexity and scale of the game. Next, we demonstrate the quality of the reward function and Nash Equilibrium policies solved by our IRL algorithm when we are given only expert demonstrations, after which we show the performance of our Nash Equilibrium algorithm when the reward function is available.

\subsection{The Chasing Game on Gridworld}

\begin{figure}[ht!]
\centering
\includegraphics[width = 0.5\textwidth]{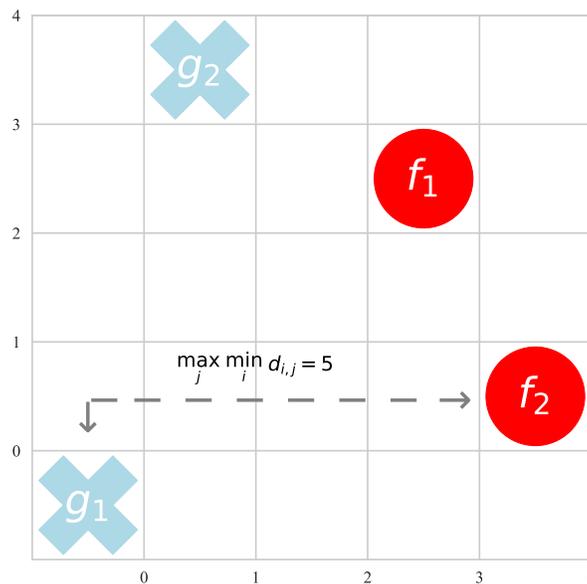}
\caption{\label{fig:game}Grid of the chasing game. We use circles to represent the predators and crosses for the preys. In this example, the distance from $g_1$ to predators is $\min_{i} d_{i,1} = 5$, while the distance from $g_2$ to predators is 3, so the distance between $f_2$ and $g_1$ determines the immediate reward at this state, and $R_{\text{chasing}}(s)=-5$.}
\end{figure}

Games on a grid have been widely used in reinforcement learning research works. In this type of games, each agent occupies one of the cells and is allowed to move to one of the neighboring cells at each step. The goal of each agent is to navigate itself to its own advantageous states, and the optimal policy depends on the relationship between the reward and location of all the agents. For instance, Abbeel \& Ng \cite{abbeel2004apprenticeship} test their single-agent IRL algorithm on a gridworld game where a small proportion of the cells have a positive reward. Reddy et al. \cite{reddy2012inverse} devise two-agent games on a small $3 \times 3$ grid where two sets of decoupled reward functions are received by the two agents respectively, driving them to avoid certain cells and move to their own rewarding cells. Lin et al. \cite{lin2017multi} simulate 1 vs 1 soccer games on $4\times 5$ or $5 \times 5$ grids where the chance of scoring, which is the reward of the game, is determined by the distance between the offensive player and the defensive player as well as the distance to the goal area. For the ``stick-together game'' used by Prasad et al. \cite{prasad2015two} the reward of each state also hinges on the distance between the two agents. We name the game chosen for our experiments as the ``chasing game'' because we flip the sign of the reward function in the stick-together game and translate the purely cooperative stick-together game into its purely competitive counterpart. Besides, we extend the 1 vs 1 game to a 2 vs 2 version, which significantly increases the complexity of the game.

As shown in Fig. \ref{fig:game}, this zero-sum stochastic game is played on a $5 \times 5$ grid. One team of predators (agent $f$) and another team of preys (agent $g$) participate in the game with two players (denoted as $f_1,f_2$ and $g_1,g_2$ respectively) in each team. For the remainder of the paper, we use circles to represent the predators and crosses for the preys in the figures, and we set the discount factor $\gamma$ as $0.9$.  At each state, any predator or prey occupies one of the cells in the grid, and each cell can contain more than one player. In terms of available actions, each player is allowed to move upward, downward, leftward, rightward, or stay, and each action is deterministically executed. At the boundary the player must stay put. At each step, the agent $f$ or $g$ simultaneously controls the two predators or preys on each team, so the game is considered to have $N_s = 25^4 = 390,625$ states with $|\mathcal{A}| = 5^2 = 25$ actions for each agent to choose from at each state. As suggested by the name of the game, the immediate return at a given state is dictated by the distances between the predators and the preys, driving the predators to pursue the preys and the preys to stay away from the predators. More formally, the distance between any predator/prey pair ($f_i$,$g_j$ with $1 \leq i,j \leq 2$) is 
$$d_{i,j} = \big|x_{f_i} - x_{g_j}\big|+\big|y_{f_i}-y_{g_j}\big|,$$
namely the L1-norm distance where state $s = (x_{f_1},y_{f_1}, x_{f_2},y_{f_2},x_{g_1},y_{g_1},x_{g_2},y_{g_2})$ is the coordinate vector of both agents. Based on this pairwise distance, the immediate reward for the predators ($f$) is
\begin{align}
R_{\text{chasing}}(s) &= - D(s) \label{chasing_reward}, \\
D(s) &= \max_{j=1,2} \min_{i=1,2} d_{i,j}, \label{chasing_distance}
\end{align}
and for the preys it is $-R_{\text{chasing}}(s)=D(s)$. In other words, the immediate reward is determined by the prey that stays the farthest from all predators. This reward function encourages cooperation and accurate allocation of tasks within a team and adds an extra layer of complexity to the game. For example, two predators chasing the same prey would result in a low return for $f$ as the other prey could conveniently run away, while the two preys that try to hide at the same corner make themselves easier to be approached simultaneously. 

In our setting there is no terminal state or ``capture'' action in the game. Therefore, even if a predator and a prey encounter each other at the same cell, the prey will not be removed from the grid, and the game will just proceed normally. Note that even though we use sampled trajectories of fixed length for our algorithms, theoretically speaking each round of the game can proceed infinitely long.

The scale of the game renders previous multi-agent IRL algorithms inefficient. Algorithms such as \cite{abbeel2004apprenticeship} and \cite{lin2017multi} are formulated under the assumption that demonstrated policies are optimal, and rely on the tabular representation of the game. When using \cite{abbeel2004apprenticeship} and \cite{lin2017multi} to solve the chasing game we have just introduced, one has to either deal with a quadratic programming problem with approximately $4\cdot 10^5$ variables and $2\cdot 10^7$ constraints, or to solve a quadratic programming problem with approximately $10^7$ variables at each iteration. In the following sections we demonstrate that our algorithms address this game with deep neural networks as model approximations and are able to yield good results.

\subsection{IRL Algorithm}

\begin{figure}%[ht!]
\centering
\includegraphics[width=0.9\textwidth]{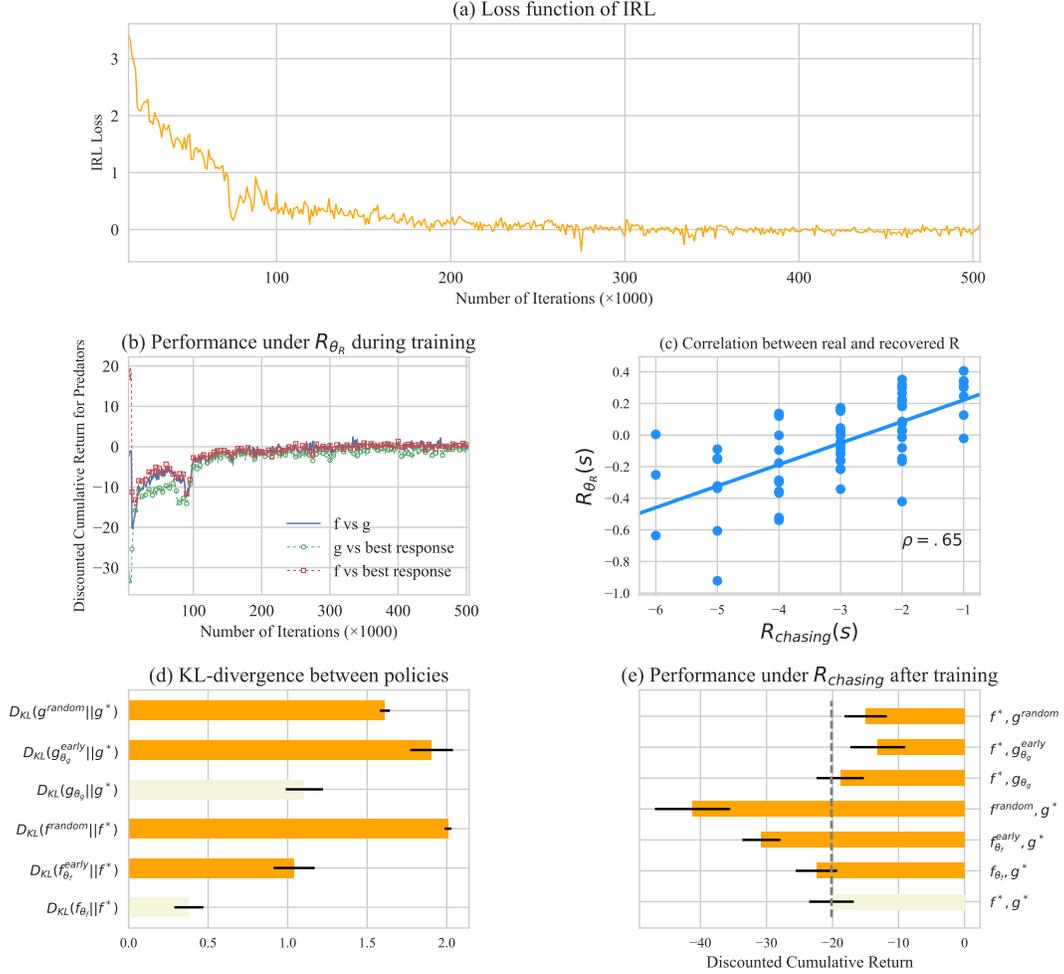}
\caption{\label{fig:IRL_train} Results of IRL training. (a) The IRL loss function, which is an estimation of objective function \eqref{obj_IRL} based on sampled trajectories, is decreasing during training. This trend indicates that $R_{\theta_R}$ gradually learns to explain the expert demonstrations in our training. (b) Performances of Nash Equilibrium policy models and their best response models during training. For the majority of the iterations, the gap of performances between $f_{\theta_f}, g_{\theta_g}$ and the best response opponent models is marginal, suggesting that $f_{\theta_f}, g_{\theta_g}$ are close enough to Nash Equilibrium policies $f^*(R_{\theta_R}), g^*(R_{\theta_R})$ during IRL training.  (c) The recovered reward function $R_{\theta_R}(s)$ demonstrates a strong correlation to $R_{\text{chasing}}(s)$ ($p<.001$). The two scales are different as reward functions are identical up to a scaling factor. (d) KL-divergence between $f_{\theta_f}$(or $g_{\theta_g}$) and $f^*(R_{\text{chasing}})$(or $g^*(R_{\text{chasing}})$). ``Early'' denotes the models at the 20,000-th iteration, while ``random'' model follows a uniform distribution on all the 5 available actions. The final results of IRL training are as expected most similar to Nash Equilibrium policies. Error bars indicate the standard errors estimated on a batch of 64 samples. (e) Performance of policy models under $R_{\text{chasing}}$. The dashed reference line represents the performance of the Nash Equilibrium model. Policies recovered by IRL training play similarly well when compared with the Nash Equilibrium policy, while ``early'' and ``random'' models exhibit much more significant performance gaps. Error bars indicate the standard deviation estimated on a batch of 64 samples. }
\end{figure}

In this section, we use the word ``iteration'' to refer to $i$ in Algorithm 1. In order to test our IRL algorithm, we use the chasing game in the setting where the immediate reward is unknown but a set of expert demonstrations is available. Details of Nash Equilibrium training are covered in Section 6.3.

The sub-optimal demonstration set $\mathcal{D}$ is generated as follows. First, under the real reward function $R_{\text{chasing}}(s)$, we use Algorithm 2 to yield Nash Equilibrium policies approximated by deep neural nets. Then we let the obtained $f^*(R_{\text{chasing}}), g^*(R_{\text{chasing}})$ to act in an $\epsilon$-greedy fashion. To be specific, we set $\epsilon=.1$ by default so that for each of the 4 players, there is $10\%$ chance that it would (1) deflect the action sampled from $f^*(R_{\text{chasing}}), g^*(R_{\text{chasing}})$ by $90^{\circ}$ or $-90^{\circ}$ if the action is not ``stay,'' or (2) randomly sample one of the 4 remaining actions if the action is ``stay,'' and $90\%$ chance it would take the original action. We thus denote the set of the demonstrated policies as $\mathcal{D}_{\epsilon=.1}$. The set consists of $64\times 500 = 32,000$ trajectories with the length of 10 steps. We believe 500 batches of trajectories are adequate since the count of states in $\mathcal{D}_{\epsilon=.1}$ is $64\times 500 \times 10 = 320,000$ and is comparable to $N_s = 390,625$. 

Aside from the $\mathcal{D}_{\epsilon=.1}$ set, we similarly generate two other sets $\mathcal{D}_{\epsilon=.05}, \mathcal{D}_{\epsilon=.2}$ so that we are able to compare the performance of the IRL algorithm under demonstration sets of various quality. Note that $\epsilon=.2$ is considered as large enough since the chance that none of the 4 players would take a deflected action is only $(1-0.2)^4=0.4096$. 

Next we describe the specifications of our models and the algorithm. The actor-critic style PPO in Algorithm 2 requires both policy models and state value function models. For both models, we use deep neural nets with a 2-layer 256-neuron structure using rectified linear units \cite{nair2010rectified} as activation functions, after which a softmax layer outputs a probability distribution on the action space in policy models (or a linear transformation layer outputs an estimated $\hat{v}(s)$ in state value models). Moreover, with the natural state vector $s$ (coordinates of players) we augment another vector $s^\prime$ which contains $x_{f_i} - x_{g_j}$ and $y_{f_i} - y_{g_j}$ for any pair $(i,j)$ with $1 \leq i,j \leq 2$ (therefore, there are 8 elements in $s^\prime$). We use the concatenated vector $(s,s^\prime)$ as the state vector. Empirically, we find that with this tailored input vector the models converge to Nash Equilibrium faster without significantly increasing the complexity of the neural networks. For results presented below we use this augmented state vector for all the models including $R_{\theta_R}(s)$. For the reward function $R_{\theta_R}(s)$ we use a 2-layer 256-neuron structure with rectified linear units as activation functions, which is followed by a linear transformation layer that outputs a scalar as the immediate reward for state $s$.

In terms of the training procedure, we set $K_R=1000$, $I_R=20$, and $\tau=3$. The learning rate for the reward function is $2.5\cdot 10^{-5}$, $T$ is set as 50, and Adam \cite{kingma2014adam} is used as optimizer. At each iteration, a batch of 64 different observations are sampled simultaneously from $\mathcal{D}$, and each of the observation provides a gradient calculated in step 11 in Algorithm 1, after which the batch of gradients are averaged to update $\theta_R$. 

Lastly, we discuss how the regularization function $\phi$ is constructed in our experiment. In \cite{lin2017multi}, Lin et al. show that the regularization term plays an important role in IRL training. This is natural since different regularizations reflect different prior knowledge about the game, and thus limit the candidates of possible reward functions to a family. In our work, we adapt their concept of the ``strong covariance'' prior to our tasks. We assume the existence of prior knowledge that the reward is related to the the distance between the predators and the preys, but it is not known that the max-min distance $D(s)$ defined in \eqref{chasing_distance} determines $R(s)$. Instead, we simply assume a negative covariance between $R_{\theta_R}(s)$ and the averaged distance 
$$\bar{D}(s) = \frac{1}{4} \sum_i \sum_j d_{i,j}. $$
Besides, to prevent the reward function from drifting too much during training, we incentivize the expected value of $R(s)$ to be close to 0 throughout training. To prevent the scale of the reward function from collapsing to 0, we incentivize the variance of $R(s)$ on the entire state space $\mathcal{S}$ to be close to a given value $\rho$. These assumptions lead to the regularization function
$$\phi(\theta_R) = c\text{ }\Big( \mathbb{E}_{s \in \mathcal{D}} cov(R_{\theta_R}(s),\bar{D}(s)) + \big| \mathbb{E}_{s \in \mathcal{D}} R_{\theta_R}(s)\big| + \big|\mathbb{E}_{s \in \mathcal{D}} var[R_{\theta_R}(s)] - \rho\big| \Big), $$
where we set $c = 0.25$ and $\rho = 5$ in our experiments. During training, we sample a batch of 64 demonstrations $\mathcal{D}^\prime \subset \mathcal{D}$ for each reward step in Algorithm 1, and calculate the regularization term based on this sample set $\mathcal{D}^\prime$ instead of $\mathcal{D}$. Though not shown in the figures, we mention that before IRL training we also initialize the reward function by training $R_{\theta_R}(s)$ for 5,000 iterations using only $\phi(\theta_R)$ above as the loss function. 

Performances of the algorithm using $\mathcal{D}_{\epsilon=.1}$ are shown in Fig. \ref{fig:IRL_train}. First of all, Fig. \ref{fig:IRL_train}(a) shows that the IRL loss function \eqref{obj_IRL} is improving during training. By IRL loss we refer to $\hat{v}^f - \hat{v}^g$ based on definitions in steps 9 and 10 in Algorithm 1, namely the objective function \eqref{obj_IRL} of our IRL algorithm (without the regularization term $\phi(\theta_R)$). This trend suggests that $R_{\theta_R}(s)$ gradually learns to explain the behaviors in the demonstration set. Meanwhile, as discussed above, the success of the IRL algorithm relies on the quality of the Nash Equilibrium policy models we maintain during IRL training. Although $\theta_R$ is being updated continuously and the Nash Equilibrium polices are expected to be changing during training, Fig. \ref{fig:IRL_train}(b) shows that the gaps between the performances of $f_{\theta_f}, g_{\theta_g}$ and their best possible performances are pretty marginal, thus indicating the good quality of both policy models. The plot depicts $v^{f_{\theta_f},g_{\theta_g}}(s_0; R_{\theta_R}), \min_{g} v^{f_{\theta_f},g}(s_0; R_{\theta_R}), \max_{f} v^{f,g_{\theta_g}}(s_0; R_{\theta_R})$, and shows that the three values are close to each other for most of the iterations during training. Regarding the property of the obtained reward function, Fig. \ref{fig:IRL_train}(c) reveals a strong correlation ($\rho=0.65, p<.001$) between $R_{\text{chasing}}(s)$ and the $R_{\theta_R}(s)$ we recovered after 500,000 iterations of training. This strong correlation indicates that the model learns that the reward of each state should be highly dependent on $D(s)$ and behaves similarly as $R_{\text{chasing}}(s)$.

To further corroborate the quality of the recovered reward and policy functions we include two more metrics. First, we compare the divergence between the IRL and Nash Equilibrium policies. As shown in Fig. \ref{fig:IRL_train}(d), we gauge the KL-divergence between the IRL and Nash Equilibrium policies and plot the estimation performed on a batch of $64$ randomly sampled states. When compared against a model that acts randomly or the ``early'' policy models obtained after 20,000 iterations, the IRL policies demonstrate behaviors that are most similar to those of the Nash Equilibrium policies.

% To further corroborate the quality of the recovered reward and policy functions we include two more metrics. First, we compare the actions taken by the IRL and Nash Equilibrium policies and count how often there is a match. As shown in Fig. \ref{fig:IRL_train}(d), we gauge the absolute difference of degrees between actions taken by the IRL and Nash Equilibrium policies and plot the distribution. The ``stay'' action is considered to be $90^{\circ}$ deflected from any other actions. The estimation is performed on $64\times 10 = 640$ randomly sampled states. When compared against a model that acts randomly or the ``early'' policy models obtained after 20,000 iterations, the IRL policies demonstrate behaviors that are most similar to those of the Nash Equilibrium policies.

\begin{figure}[ht!]
\centering
\includegraphics[width=1\textwidth]{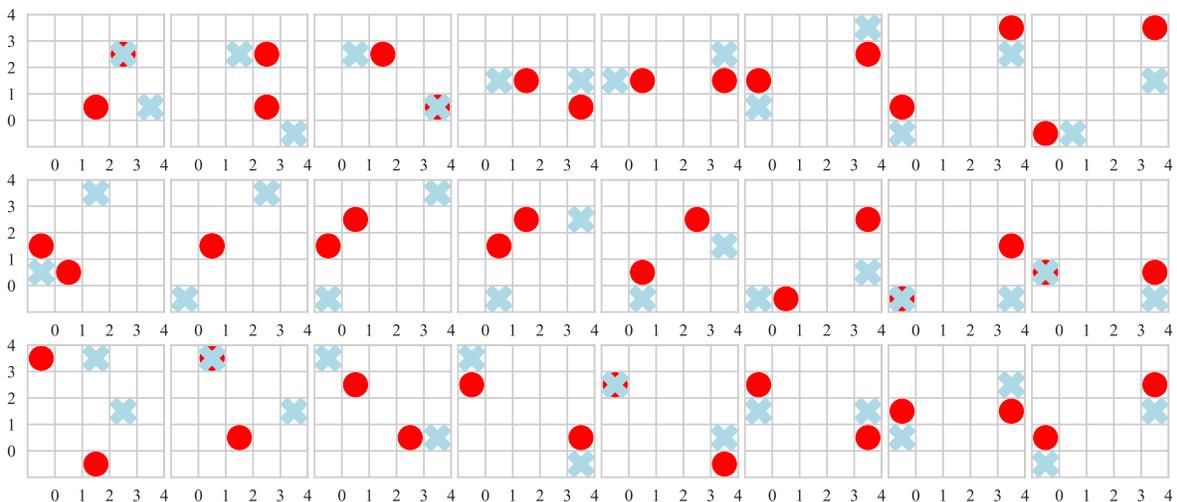}
\caption{\label{fig:IRL_traj}Trajectories generated by policy models obtained in the IRL algorithm. Each row presents a different 8-step trajectory.}
\end{figure}

\begin{figure}[ht!]
\centering
\includegraphics[width=1\textwidth]{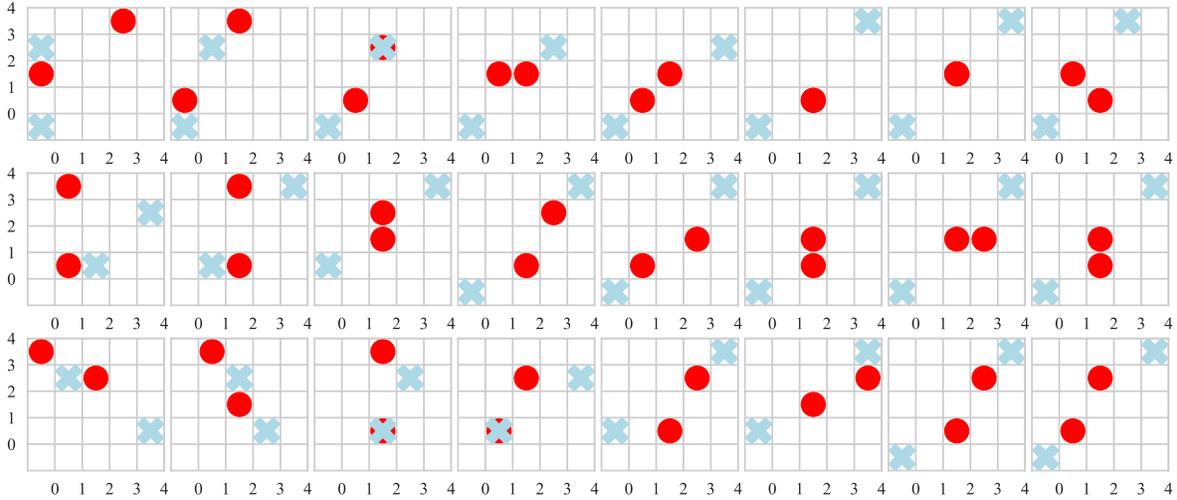}
\caption{\label{fig:IRL_traj_early}Trajectories generated by policy models obtained at the 20,000-th iteration. Each row presents a different 8-step trajectory. Clearly, these actions are not driven by $R_{\text{chasing}}(s) = D(s)$ and are different from the ones in Fig. \ref{fig:IRL_traj}.}
\end{figure}

A more direct measurement is to plug IRL policies back into the chasing game and evaluate their performances when competing against the Nash Equilibrium policies. In Fig. \ref{fig:IRL_train}(e) we depict the performances of the IRL and Nash Equilibrium polices estimated in 64 rounds of games. The policies $f_{\theta_f}, g_{\theta_g}$ obtained after 500,000 iterations of IRL training demonstrate performances that are relatively close to those of Nash Equilibrium policies.

% Given the discussion in Section $4$, we know that the success of our IRL algorithms relies on the qualities of $f_{\theta_f}$ and $g_{\theta_g}$ models. The further they diverge from the Nash Equilibrium strategies under the current $R_{\theta_R}$, the worse the estimation at step 9 and 10 would be in Algorithm 1.  Since the reward step is performed very occasionally, it is natural that $f_{\theta_f}, g_{\theta_g}$ are able to notice and adjust to the latest changes on $R_{\theta_R}$. 

% Regarding the training on the reward function and its quality, Fig. \ref{fig:IRL_train}(b) shows that the loss function in IRL is improving after the first few initialization iterations. By IRL loss we refer to $\hat{v}^f - \hat{v}^g$ based on definition at steps 9 and 10 in Algorithm 1, namely the objective function \eqref{obj_IRL} of our IRL algorithm (Note that the regularization term $\phi(\theta_R)$ is not included). Besides, Fig. \ref{fig:IRL_train}(c) reveals a negative correlation ($\rho=-0.65$) between $D(s)$ and $R(s)$ where a batch of 64 $s \in \mathcal{D}$ are used as samples. Since we do not coerce the range of $R_{\theta_R}(s)$ to be the same as \eqref{chasing_reward}, we care more about this correlation rather than the exact value and scale of $R(s)$.
\begin{figure}[ht!]
\centering
\includegraphics[width=1\textwidth]{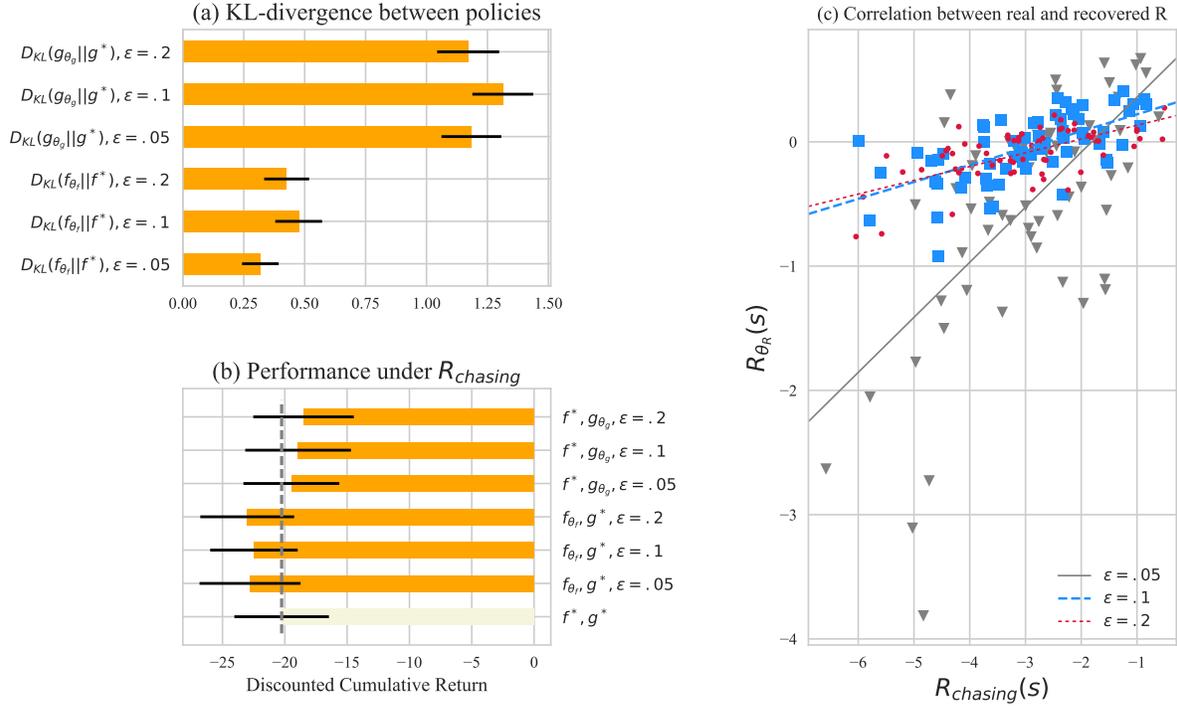}
\caption{\label{fig:IRL_traj_multipleEpsilon}Comparison of IRL performances under $\mathcal{D}_{\epsilon}$ with different $\epsilon$ values. (a) KL-divergence between the IRL and Nash Equilibrium policies are similar under different $\mathcal{D}_{\epsilon}$. Error bars indicate the standard errors estimated on a batch of 64 samples. (b) Using different $\mathcal{D}_{\epsilon}$ we yield IRL policies that perform similarly under the original $R_{\text{chasing}}(s)$. The performances are measured in the same way as in Fig. \ref{fig:IRL_train}(e). Error bars indicate the standard deviation estimated on a batch of 64 rounds of games. (c) Range of the recovered $R_{\theta_R}(s)$ under different $\mathcal{D}_{\epsilon}$. The larger the $\epsilon$ is, the smaller the range of $R_{\theta_R}(s)$. For readability of the figure a jitter is added to the $x$-coordinate of each scatter point.   }
\end{figure}

In Fig. \ref{fig:IRL_traj} we demonstrate the behaviors of IRL policies based on random trajectories generated by $f_{\theta_f},g_{\theta_g}$. Despite occasional mistakes (for example, in the first row of Fig. \ref{fig:IRL_traj} one of the predators chose not to move), the two predators are pursuing the preys in a coordinated way, and the preys are actively keeping a distance from the predators.

% \begin{figure}[ht!]
% \centering
% \includegraphics[width=1\textwidth,trim={0 0 78cm 0}]{IRL_multipleEpsilon.eps}
% \caption{\label{fig:IRL_traj_multipleEpsilon}Comparison of IRL performances under $\mathcal{D}_{\epsilon}$ with different $\epsilon$ values. (a) The divergence of IRL policies and Nash Equilibrium policies are similar under different $\mathcal{D}_{\epsilon}$. Divergence of actions is measured in the same way as in Fig. \ref{fig:IRL_train}(d). (b) Using different $\mathcal{D}_{\epsilon}$ we yield IRL policies that perform similarly under the original $R_{\text{chasing}}(s)$. The performances are measured in the same way as in Fig. \ref{fig:IRL_train}(e). Error bars indicate the standard deviation estimated on a batch of 64 rounds of games. (c) Range of the recovered $R_{\theta_R}(s)$ under different $\mathcal{D}_{\epsilon}$. The larger the $\epsilon$ is, the smaller the range of $R_{\theta_R}(s)$. Shades indicate the $95\%$ CI. For readability of the figure a jitter is added to the $x$-coordinate of each scatter point.   }
% \end{figure}

There remains a concern on whether the prior knowledge in our regularization function is too strong. If so, we should have approximated $R_{\text{chasing}}(s)$ decently well from the beginning of our training. We clear this doubt by inspecting the Nash Equilibrium policies early in the algorithm. In Fig. \ref{fig:IRL_traj_early}, we show trajectories generated by policy models at the 20,000-th iteration. Note that we also use models obtained at the 20,000-th iteration as the ``early'' models in Fig. \ref{fig:IRL_train} because at the 20,000-th iteration the Nash Equilibrium polices are of good qualities already (shown in Fig. \ref{fig:IRL_train}(b)) while IRL training has just begun (shown in Fig. \ref{fig:IRL_train}(a)). Obviously, for trajectories in Fig. \ref{fig:IRL_traj_early} both agents act remarkably different from policies shown in Fig. \ref{fig:IRL_traj}. To be specific, in Fig. \ref{fig:IRL_traj_early}  the preys try to move to and stay at two corners on the diagonal of the grid, while the predators try to stay on the diagonal of the two preys. This is not surprising since in $\phi(\theta_R)$ we are encouraging the average distance $\bar{D}(s)$ and $R(s)$ to be correlated instead of the max-min distance $D(s)$ and $R(s)$, and the behaviors of predators/preys serve to minimize/maximize $\bar{D}(s)$. Therefore, we confidently draw the conclusion that the policies and the reward function are recovered by our IRL algorithm because of the correctly proposed objective function that minimizes the performance gap rather than the prior knowledge provided by the regularization terms.

A final comparison between models recovered by the IRL algorithm using $\mathcal{D}_{\epsilon=.05}$, $\mathcal{D}_{\epsilon=.1}$, and $\mathcal{D}_{\epsilon=.2}$ illustrates the robustness of the performance of the IRL algorithm despite the variation of the qualities in expert demonstrations. In Fig. \ref{fig:IRL_traj_multipleEpsilon}(a) and (b) we show that IRL policies produced under different $\mathcal{D}_{\epsilon}$ behave similarly when compared against Nash Equilibrium policies, and demonstrate nearly the same performances in the original chasing game. In Fig. \ref{fig:IRL_traj_multipleEpsilon}(c), we show that the quality of the demonstration set affects the range of the reward function. Even though a similarly strong correlation ($r\approx.65$) is produced for each $\mathcal{D}_{\epsilon}$, the scale of $R_{\theta_R}(s)$ decreases when $\epsilon$ goes up. The larger the $\epsilon$ is, the less distinctive the rewards of different states are in the reward function inferred by the IRL algorithm. This also explains why the different policies functions we recover perform similarly as shown in Fig. \ref{fig:IRL_traj_multipleEpsilon}(a) and (b), since rescaling the reward function does not strongly affect the agents' preferences for different states, and thus the optimal policies remain largely unchanged. This is ideal as the success of IRL training is not critically influenced by the quality of the available demonstration set, and from sub-optimal demonstrations of distinct qualities the algorithm recovers policies that perform very well.

\subsubsection{Key Findings}

The proposed IRL algorithm aims to minimize the performance gap between the Nash Equilibrium policies and the sub-optimal policies demonstrated in $\mathcal{D}$. Under appropriate regularization, the gap decreases to a relatively marginal level within the first 100,000 iterations of training, and we successfully recover the reward function and the optimal policies for the chasing game after 500,000 iterations of training. The recovered reward function $R_{\theta_R}(s)$ exhibits a strong correlation to $R_{\text[chasing]}(s)$. The recovered policies not only behave similarly when compared with the Nash Equilibrium policies, but also demonstrate good performance when competing against them in the chasing game. We also observe that the quality of the demonstration set affects the scale of the learned $R_{\theta_R}(s)$, while the performances of the recovered policies remain largely unchanged, which suggests the robustness of IRL training despite variation in demonstrated sub-optimal policies. 

\subsubsection{Comparison with existing IRL algorithms}

Our IRL algorithm demonstrated is designed and tailored for two specific goals: to take sub-optimality of expert demonstrations into account, and to cope with zero-sum games of large scales. We next illustrate the superior performance of our algorithm in the chasing game regardless of the quality of the demonstration set. The Bayesian-IRL \cite{lin2017multi} (BIRL) algorithm and Decentralized-IRL \cite{reddy2012inverse} (DIRL) are selected as benchmark IRL algorithms since they are the only ones that solve competitive multi-agent IRL tasks to the best of our knowledge. Both algorithms need modifications since deep neural nets should be used as model approximations and the IRL training should proceed efficiently for large-scale games that cannot be solved by tabular approaches with enumeration of states or actions. We next provide the details.

% To make sure we are performing a fair comparison, the number of training iteration for each algorithm is set to match the runtime of all 3 algorithms. Specifications for both BIRL and DIRL algorithms in our experiments are given below. 

The BIRL algorithm for zero-sum stochastic games formulates a quadratic programming problem with constraints that require each demonstrated action to be the optimal one, and the objective function represents the posterior of the current reward function given a Bayesian prior of reasonable reward functions. Two problems arise when implementing the BIRL algorithm to solve the chasing game. First, enumerating and imposing all the constraints is not tractable (there are $2\cdot N_{s}\cdot |\mathcal{A}| = 19,531,250$ constraints in the current version of the game). Therefore, for each iteration in our training, we sample a tuple $(s,a^{E,f},a^{E,g})$ from $\mathcal{D}$, and randomly choose actions $(a^f,a^g)$ from $\mathcal{A}^f \times \mathcal{A}^g$. In an iteration we consider only the constraints on the sampled state-action pairs. By adding Lagrangians into the objective function to encourage the constraints, we charge a penalty whenever the demonstrated actions $a^{E,f},a^{E,g}$ are performing worse than $a^f, a^g$. 

Another issue is that the BIRL algorithm requires explicit models for expert policies. The aforementioned constraints in the BIRL algorithm are equivalent to $Q^{f^{E},g^{E}}_f(s,a^{E,f}) \geq Q^{f^{E},g^{E}}_f(s,a^{f}) $ and $Q^{f^{E},g^{E}}_g(s,a^{E,g}) \geq Q^{f^{E},g^{E}}_g(s,a^{g})$. Evaluation of the $Q$-functions are feasible only if $f^{E},g^{E}$ and the corresponding state transition matrix are available and can be stored in a computer's memory. In \cite{lin2017multi} the expert policies are statistically recovered since sufficient demonstrations are available for a significantly smaller game, which is impossible for large games. Instead of appealing to imitation learning to yield expert model approximations, we conduct a two-phase training that does not rely on expert policy models. We find the optimal state value function first, and then use the state-value function and one-step transition probability (which can be approximated by sampling tuples from $\mathcal{D}$) to recover $R$. To be specific, the BIRL algorithm is based on the equality $\bm{V} = (\bm{I} - \gamma \bm{P})^{-1}\bm{R}$ where $\bm{V} = \big(V(s)\big)_{s\in\mathcal{S}}$ is the vector exhibiting the value for each state, $\bm{R} = \big(R(s)\big)_{s\in\mathcal{S}}$ is the vector for the reward at each state, and $\bm{P}$ is the state transition matrix under expert policies. To infer $\bm{V}$ from $\bm{R}$, the inversion of $(\bm{I} - \gamma \bm{P})^{-1}$ necessitates expert policy models that can act in all states (including those not demonstrated in $\mathcal{D}$) and generate infinitely long trajectories. Instead, if the algorithm first finds state value functions $V(s)$ instead of $R(s)$ and uses $\bm{R} = (\bm{I} - \gamma \bm{P})\bm{V}$ to recover $R$, then only one-step transitions that can be sampled directly from $\mathcal{D}$ are needed. Therefore, in our implementation the first phase of training uses the BIRL algorithm to solve for $v^{f^{E},g^{E}}_{\theta_{V}}(s)$, the vector representation of which is the vector $\bm{V}$ above. The objective function is equal to the Lagrangian terms plus the same regularization term $\phi(\theta_R)$ used for our IRL algorithm (now viewed as prior of $R$ in BIRL). In the second phase, we sample a state $s$ and corresponding expert actions from $\mathcal{D}$, get the following state $s^\prime$ under the known transition function, and train $\theta_R$ to minimize the squared loss between $R_{\theta_R}(s)$ and $v^{f^{E},g^{E}}_{\theta_{V}}(s) - \gamma v^{f^{E},g^{E}}_{\theta_{V}}(s^\prime)$. Note that the objective function in the $R$-phase is also regularized by the same $\phi(\theta_R)$ in the $V$-phase, because in our experiments we have observed a drastic deterioration of performances if the regularization term is not used for both phases. The model $v^{f^{E},g^{E}}_{\theta_{V}}(s)$ and reward $R_{\theta_R}(s)$ are parametrized similarly as specified in Section 6.2. Lastly, since the BIRL algorithm returns only a reward function, we use the proposed Nash Equilibrium algorithm to solve for $f^*(R_{\theta_R}), g^*(R_{\theta_R})$ after two-phase training.

The DIRL algorithm also assumes the optimality of expert policies under the unknown reward function. The algorithm alternates between a $\pi$ step and an $R$ step. In the $k$-th iteration of training, the algorithm first enters the $\pi$ step that solves for the Nash Equilibrium policies $(f_k,g_k)$ under current $R_{\theta_R}$. The policies $(f_k,g_k)$ are added into a policy set $\Pi$. Then in the $R$ step, the algorithm finds $R_{\theta_R}$ that maximizes $\frac{1}{k}\sum_{j=1}^{k}\sum_{s\in\mathcal{S}}p\Big(v^{f^{E},g^{E}}(s) - v^{f_j,g^{E}}(s) \Big) + p\Big( v^{f^{E},g_j}(s) - v^{f^{E},g^{E}}(s) \Big) - \phi(\theta_R)$, where $p(x) = \max(x,0) + 2\cdot\min(x,0).$ This objective function encourages $R$ to favor $f^E,g^E$ when competing against any policies in $\Pi$, which is aligned with the optimality assumption that $(f^E,g^E)$ are indeed Nash Equilibrium of the game. 

To alleviate the overhead of storing and calling all the policies in $\Pi$, in the $R$ step of our deep implementation we maximize a slightly different objective function $\mathbb{E}_{(f_j,g_j) \sim \Pi} \mathbb{E}_{(s,\_,\_)\sim\mathcal{D}}\bigg[p\Big(v^{f^{E},g^{E}}(s) - v^{f_j,g^{E}}(s) \Big) + p\Big( v^{f^{E},g_j}(s) - v^{f^{E},g^{E}}(s) \Big) - \phi(\theta_R)\bigg]$. Thus, in each training iteration we only sample one pair of $(f_j,g_j)$ from $\Pi$ and pit them against expert policies, so the expectation of this new objective function remains unchanged when compared with the original one. Again, models of $f^E,g^E$ are still required to evaluate the state value functions. Here we adopt the treatment of the original DIRL work \cite{reddy2012inverse} and our IRL algorithm; we let $f^E,g^E$ to act at the first step of each trajectory by sampling from $\mathcal{D}$, then we use the latest $(f_k,g_k)$ to act for all the following steps to generate the full trajectory.

To make sure we are performing a fair comparison, the number of training iteration for each algorithm is set to match the runtime of all 3 algorithms. For both phases in the algorithm, training lasts for 500,000 iterations. For the Lagragians in BIRL training, we use a fixed coefficient for all constraints instead of a unique and dynamically updated coefficient for each one, which would theoretically require another neural network models for $\lambda(s,a)$. Besides, the fixed coefficient is set to be $1$ since we can change the weights in $\phi(\theta_R)$ instead, which is similar to the original treatment in \cite{lin2017multi}. We set the weight coefficient $c$ in $\phi(\theta_R)$ to be $.25$ to match up with the one in our algorithm. The DIRL algorithm is computationally demanding largely due to the time spent on solving for Nash Equilibrium at each iteration of training. To control the runtime of the DIRL algorithm within a comparable range of the other IRL methodologies in our experiment, we perform 10 iterations with 50,000 training iterations for both the $\pi$ and $R$ steps in each iteration. For both algorithms, policies and reward function models are parametrized similarly as specified in Section 6.2. The learning rate is set to be $10^{-4}$, weight $c$ of regularization term is set as $.25$ and Adam \cite{kingma2014adam} is used as optimizer. We mention that, even under such a specification, DIRL training still more than tripled the runtime of the other algorithms in our experiments. 

\begin{table}[t]
  \caption{Correlations between recovered reward functions and $R_{\text{chasing}}$}
  \label{corr_irl_benchmark}
  \centering
  
\begin{tabular}{l*{3}{c}r}
IRL Algorithm               & $\epsilon = .05$ & $\epsilon = .1$ & $\epsilon = .2$  \\
\hline
Algorithm 1 & .65 & .68 & .66    \\
BIRL            & .28 & -.02 & .12    \\
DIRL           & -.31 & -.15 & .11    \\
\end{tabular}

\end{table}

\begin{table}[t]
  \caption{Performance deterioration of recovered policies under $R_{\text{chasing}}$ (*:Nash Eq.; A:Algorithm 1; B:BIRL; D:DIRL)}
  \label{performance_irl_benchmark}
  \centering
  
\begin{tabular}{l*{6}{c}r}
\textbf{$\mathcal{D}_{\epsilon}$ }             & $f^{\text{A}}$ & $f^{\text{B}}$ & $f^{\text{D}}$ & $g^{\text{A}}$  & $g^{\text{B}}$ & $g^{\text{D}}$  \\
\hline
\textbf{$\epsilon = .05$ }			& 11.8\% & 24.1\% & 197.0\% & 4.4\%  & 33.5\% & 38.9\%  \\
\textbf{$\epsilon = .1$		}		& 10.3\% & 44.3\% & 100.0\% & 6.9\%  & 33.5\% & 41.9\%  \\
\textbf{$\epsilon = .2$	}			& 13.3\% & 68.5\% & 200.1\% & 9.3\%  & 33.0\% & 40.9\%  \\
\end{tabular}
  
% \begin{tabular}{l*{6}{c}r}
% \textbf{$\mathcal{D}_{\epsilon}$ }             & $f^*,g^*$ & $f^{\text{A}},g^*$ & $f^{\text{B}},g^*$ & $f^{\text{D}},g^*$ & $f^*,g^{\text{A}}$  & $f^*,g^{\text{B}}$ & $f^*,g^{\text{D}}$  \\
% \hline
% \textbf{$\epsilon = .05$ }			& -20.3 & -22.7 & -25.2 & -60.3 & -19.4  & -13.5 & -12.4  \\
% \textbf{$\epsilon = .1$		}		& -20.3 & -22.4 & -29.3 & -40.6 & -18.9  & -13.5 & -11.8  \\
% \textbf{$\epsilon = .2$	}			& -20.3 & -23.0 & -34.2 & -61.1 & -18.4  & -13.6 & -12.0  \\
% \end{tabular}

\end{table}

We summarize the result of experiments in Table 1 and 2. Under the same regularization terms, only our IRL algorithm finds a reward function that bears reasonably high correlation with $R_{\text{chasing}}(s)$. We also plug the solved IRL policies back into the original chasing game and compete against the Nash Equilibrium policies, and measure the gap between their performances and $v^{f^*,g^*}$ to evaluate how much the performance of the recovered policies deteriorate. As shown in Table 2, only our IRL algorithm recovers policies of good quality, and the performance is not largely affected by the demonstration set we use. The issues with the BIRL algorithm are the requirement of accurate expert policy models and the strict optimality constraints of expert actions, whereas the number of reference policies in $\Pi$ is likely to be highly critical to the success of the DIRL algorithm. In conclusion, our IRL algorithm overcomes the issues in the benchmark algorithms, and outperforms them significantly when all the algorithms are implemented and utilized in the same setting.

\subsection{Nash Equilibrium Algorithm}

\begin{figure}[ht!]
\centering
\includegraphics[width=1\textwidth]{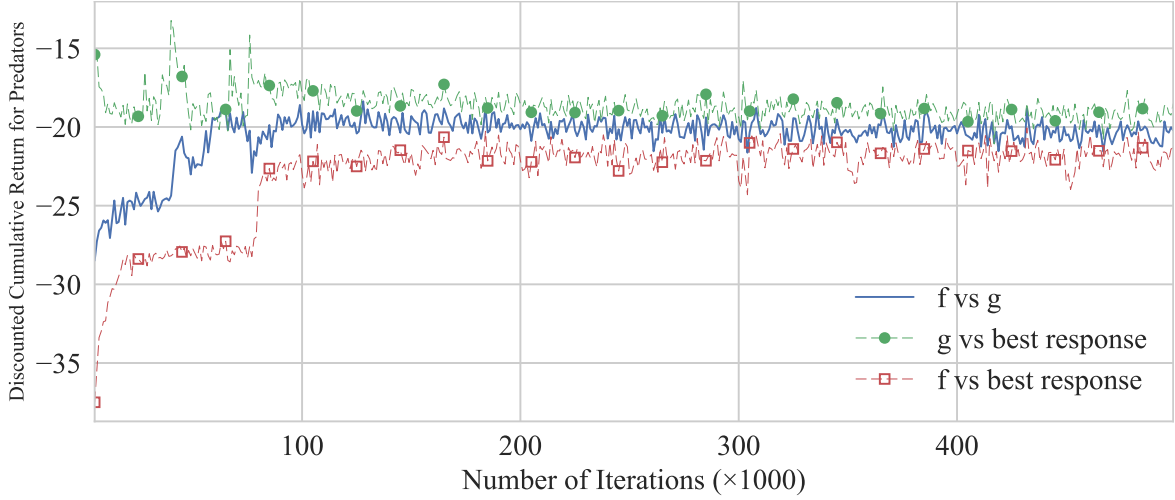}
\caption{\label{fig:nasheq_train}Performance of the proposed Nash Equilibrium algorithm in the chasing game}
\end{figure}

\begin{figure}[ht!]
\centering
\includegraphics[width=1\textwidth]{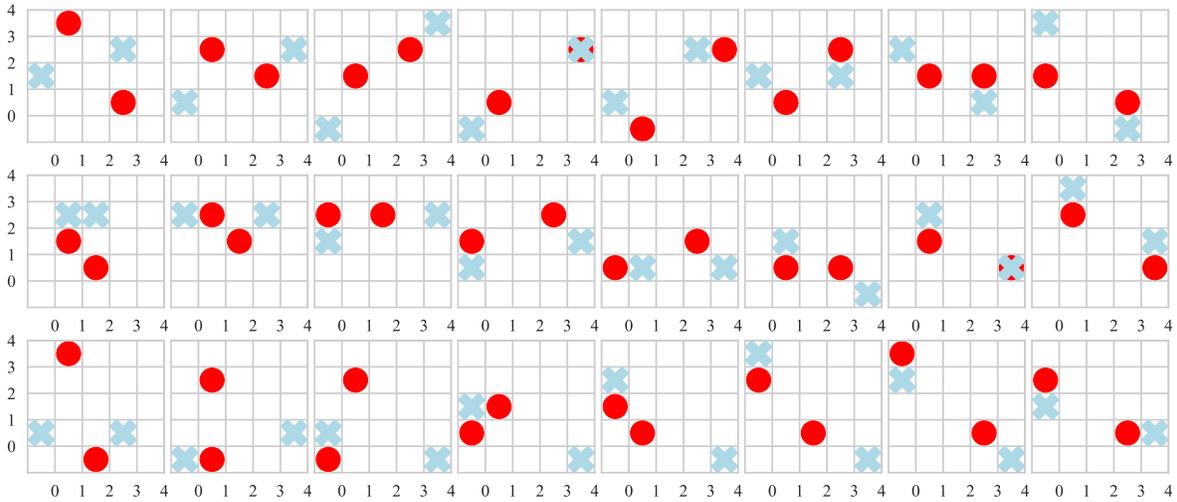}
\caption{\label{fig:nasheq_traj}Trajectories generated by the trained models in the chasing game. Each row presents a different 8-step trajectory.}
\end{figure}

The Nash Equilibrium algorithm assumes that the reward function is available. We herein showcase the algorithm's performance under $R_{\text{chasing}}(s)$. As noted before, Algorithm 2 is meant for finding $f^*(R)$, and a separate training procedure is carried out to solve for $g^*(R)$. We only discuss in detail training of $f^*(R)$ because the same hyper-parameters and model structures are used for training $g^*(R)$. 

In this section, we use the word ``iteration'' to refer to $i$ in Algorithm 2. The following hyper parameters are used in Algorithm 2 for the experiments in this section and those already presented in Section 6.2. When performing PPO style training, we set horizon length $T$ as 10, and refresh frequency $K_{\text{refresh}}$ as 10. Parameter $\lambda$ for eligibility traces is set as $0.9$. Regarding the adversarial training, we set $K_{\text{cycle}}$ as 100 and $K_g$ as 90, under which the frequency of the $g$ and $f$ step is $9:1$. As mentioned above, a batch of 64 agents (parametrized by current $\theta_f^{\text{target}}, \theta_g^{\text{target}}$) act simultaneously from independently initialized starting states. The 64 trajectories are used together for the stochastic gradient descent steps in the $g$ or $f$ step in Algorithm 2. Meanwhile, inspired by \cite{arjovsky2017wasserstein}, for the first 5,000 iterations of every 50,000 iterations we perform the $g$ step only. This is to ensure the quality of $g_{\theta_g}$, which is expected to be the best possible response to current $f_{\theta_f}$ during training. In our experiment, we used Adam \cite{kingma2014adam} as the optimizer due to its superior performance when training deep models. The learning rate for best response models is set as $3\cdot 10^{-4}$, while the learning rate for the Nash Equilibrium policies $f_{\theta_f}, g_{\theta_g}$ is $10^{-4}$. This configuration makes sure that changes in the latest $f_{\theta_f}$ are slow enough for the best response opponent model to adjust itself. Again, we use Adam to perform stochastic optimization on our models.

% Specifically, in order to employ the actor-critic style PPO algorithm, both policy models and state value function models are required. For both models, we use deep neural nets with a 2-layer 256-neuron structure using rectified linear units \cite{nair2010rectified} as activation functions, after which a softmax layer (or a linear multiplication) outputs a probability distribution on the action space for policy models (or outputs an estimated $\hat{v}(s)$ for state value functions). Moreover, rather than feeding the state vector $s$ (coordinates of players) directly into the model, we engineer the input and incorporate another vector $s^\prime$ which contains $x_{f_i} - x_{g_j}$ and $y_{f_i} - y_{g_j}$ for any pair $(i,j)$ with $1 \leq i,j \leq 2$ (therefore, there are 8 elements in $s\prime$). We feed the concatenated vector $(s,s^\prime)$ into the model. Empirically, we find with this tailored input vector the models converge to Nash Equilibrium faster without increasing the complexity of the neural networks, and for the results presented below we use this tailored input for all the models. 

Fig. \ref{fig:nasheq_train} shows the evolution of the performances of policy models. The plot depicts $v^{f_{\theta_f},g_{\theta_g}}(s_0; R_{\text{chasing}})$, $\min_{g} v^{f_{\theta_f},g}(s_0; R_{\text{chasing}})$, $\max_{f} v^{f,g_{\theta_g}}(s_0; R_{\text{chasing}})$. Recall that if $f_{\theta_f}$ and $g_{\theta_g}$ manage to reach the Nash Equilibrium, the three series should converge perfectly. As the figure shows, the adversarial training on $f_{\theta_f}$ and $g_{\theta_g}$ succeed in decreasing the gap to a pretty marginal level, suggesting that $f_{\theta_f}$ and $g_{\theta_g}$ should be close enough to the Nash Equilibrium policies and there is little room to further improve their performances.

To further corroborate the quality of the learned policies, in Fig. \ref{fig:nasheq_traj} we plot trajectories showing behaviors of the obtained $f_{\theta_f}$ and $g_{\theta_g}$. As demonstrated in the figure, both the predators and the preys understand their goals in the game. The predators allocate the tasks so that they are not chasing the same prey and ignoring the other one, while the strategy of the preys is that they run away actively and try not to stay at the same cell, thus making it harder for the predators to pursue both of them.

\subsubsection{Key Findings}

By adopting the framework of adversarial training and alternating between the Nash Equilibrium policies and the best response policies, the proposed Nash Equilibrium algorithm successfully solves the chasing game. The performance of the Nash Equilibrium policy reaches a decent level within the first 100,000 iterations of training. Both the predator and the prey models demonstrate a strategy that is driven by $R_{\text{chasing}}(s)$, and the two players on the same team learn to cooperate with each other. 

\subsubsection{Comparison with existing Nash Equilibrium algorithm}

To further demonstrate the superiority of the proposed Nash Equilibrium algorithm particularly for large games, we reformulate the quadratic programming problem proposed on page 125 in \cite{filar2012competitive}, which inspires the gradient descent algorithm to solve for Nash Equilibrium proposed in \cite{prasad2015study}. We select only this algorithm as the benchmark in this section, since the algorithm in \cite{prasad2015two}, due to its formulation, was observed to provide a zero gradient to policies when training in zero-sum games, and we do not find an easy solution to apply the algorithm in  \cite{akchurina2009multiagent} to large games using deep neural nets as model approximations.

Here we illustrate the deep implementation we used for the benchmark algorithm in the experiment. The algorithm maintains both policy models $f,g$ and models for bounds of state value functions $v^f(s),v^g(s)$. The softmax layers in policy models guarantee $f(a|s),g(a|s)$ to be well-defined distributions. Therefore, the remaining constraints are 
$$R(s) + \gamma\mathbb{E}_{a^g\sim g(a|s), s^\prime \sim p(s^\prime|s,a^f,a^g) }v^f(s^\prime)\leq v^f(s) \text{ for any } s, a^f,$$
$$-R(s) + \gamma\mathbb{E}_{a^f\sim f(a|s), s^\prime \sim p(s^\prime|s,a^f,a^g) }v^g(s^\prime)\leq v^g(s) \text{ for any } s, a^g,$$ 
and the objective is to minimize $\mathbb{E}_s[v^f(s)+v^g(s)]$. The constraints are implemented as Lagrangians with a coefficient $\lambda$. Similarly to the implementation of the benchmark BIRL algorithm, we do not enumerate all the constraints, but only sample $s$ from $S$ and $a^f,a^g$ from $f(a|s),g(a|s)$ at each iteration of training. Whenever a constraint on the sampled state-action pair is violated, a penalty is added to the objective function that motivates $f,g$ to avoid taking sub-optimal actions. To evaluate the expectation in each constraint, we sample 5 trajectories for each constraint. Training lasts for 500,000 iterations. Both policies and value models are parametrized as in our Nash Equilibrium algorithm. Theoretically speaking, for each state-action pair, the corresponding constraint should have its own $\lambda$ (which necessitates an extra neural network model) that would be constantly updated in each iteration. Since we are not enumerating the constraints, we set $\lambda$ to be a fixed value throughout training, and conduct grid search on the optimal value of $\lambda$. According to our experiments, performance of the algorithm is not very sensitive to the value $\lambda$, and we use $\lambda = 10$ for the results shown in this section since it appears to be the optimal value in our experiments.

We summarize the results in Table 3. In our experiments we compared the performances of both algorithms in the original chasing game (with a $5\times 5$ grid) and a larger game (with a $10\times 10$ grid). The values are the average of a batch of 64 trajectories with randomly initialized starting states. We also show the performances under random policies as a reference. For the game on the $5\times 5$ grid, we see that our algorithm yields a better policy for predators, while preys of both algorithms perform similarly. For the game on the $10\times 10$ grid, the benchmark algorithm learned a much worse policy for predators (the improvement of our algorithm is $\frac{-87.4 - (-44.7)}{-87.4}=48.6\%$ from the benchmark algorithm), and the policies for preys learned by our algorithm are $\frac{-44.7 - (-42.2)}{-42.2} = 5.9\%$ better than that of the benchmark algorithm. Overall, our adversarial training algorithm for Nash Equilibriums in zero-sum stochastic games shows a significantly improved performance against the benchmark algorithm, especially for larger cases.

\begin{table}[t]
  \caption{Performances of solved Nash Equilibrium policies (A:Algorithm 2; B:Benchmark; R:Random)}
  \label{performance_nasheq_benchmark}
  \centering
  
\begin{tabular}{l*{6}{c}r}
Grid Size             & $f^{\text{A}},g^{\text{A}}$ & $f^{\text{B}},g^{\text{A}}$ & $f^{\text{R}},g^{\text{A}}$ & $f^{\text{A}},g^{\text{B}}$ & $f^{\text{A}},g^{\text{R}}$   \\
\hline
$5\times5$  & -20.3 & -21.2 & -41.1 & -20.0 & -14.9    \\
$10\times10$& -44.7 & -87.4 & -94.2 & -42.2 & -31.8		\\

\end{tabular}

\end{table}

\section{Conclusion and Future Work}

IRL tasks differ significantly from traditional reinforcement learning tasks. In IRL settings, we are given a set of expert demonstrations to infer the underlying reward function that drives the observed behaviors. For competitive multi-agent IRL tasks, existing methods assume the optimality of expert demonstrations, and the two agents involved in a zero-sum stochastic game are decoupled in those algorithms. In this paper, we propose a new framework for competitive multi-agent IRL tasks that takes sub-optimality of expert demonstrations into account.

We propose an IRL algorithm with the objective to explicitly minimize the performance gap between expert policies and Nash Equilibrium strategies. In order to solve for a Nash Equilibrium strategy, we also propose an adversarial training algorithm, and show its theoretical appeal in the non-existence of local optimum in the objective function. For our experiments, neural networks are used as model approximations so that the algorithm recovers both the reward and the policy functions in a game that is too large for existing competitive multi-agent IRL methodologies.

The work concerns discounted stochastic games where the state information is completely public to both agents. For future research, a natural extension to the current work is to adapt the framework to partially observable MDPs, Nevertheless, as mentioned in \cite{reddy2012inverse} finding Nash Equilibriums becomes much more challenging than in our case. Meanwhile, another natural generalization is to explore how our gradient-based Nash Equilibrium algorithm can be applied to general-sum games.

Our theoretical analysis of Algorithm 2 suggests that the gradient to improve the Nash Equilibrium policy models should be evaluated against all the best response models. Therefore, it is an interesting direction to see how a multitude of best response opponent models can help improve the performance of the Nash Equilibrium algorithm. Besides, scaling up the Nash Equilibrium algorithm for even larger games might entail improvements on the efficiency and stability of our adversarial training. Aside from existing works such as \cite{foerster2017learning}, it is worth exploring how a better estimated gradient can be provided to policy models in the algorithm, and how the policy models realize and adjust to the changes in their opponents.

\bibliographystyle{unsrt}
\bibliography{bib_appendix}

\end{document}